\documentclass[letterpaper]{article} 
\usepackage{aaai2026_arxiv}  
\usepackage{times}  
\usepackage{helvet}  
\usepackage{courier}  
\usepackage[hyphens]{url}  
\usepackage{graphicx} 
\usepackage{natbib}  
\usepackage{caption} 
\usepackage[dvipsnames]{xcolor}

\usepackage{kotex}
\usepackage{lipsum}
\usepackage{pifont}
\newcommand{\cmark}{\ding{51}}%
\newcommand{\xmark}{\ding{55}}%
\usepackage{mathtools}

\usepackage{array}
\newcolumntype{P}[1]{>{\centering\arraybackslash}p{#1}}
\newcolumntype{M}[1]{>{\centering\arraybackslash}m{#1}}
\newcolumntype{L}[1]{>{\hspace{0.5em}\raggedright\arraybackslash}m{#1}}
\newcolumntype{R}[1]{>{\raggedleft\arraybackslash}m{#1}}

\usepackage{xspace}

\makeatletter
\DeclareRobustCommand\onedot{\futurelet\@let@token\@onedot}
\def\@onedot{\ifx\@let@token.\else.\null\fi\xspace}

\def\eg{\emph{e.g}\onedot} 
\def\ie{\emph{i.e}\onedot} 
 
 \def\vs{\emph{vs}\onedot}

\makeatother

\usepackage[colorlinks=true, linkcolor=blue, citecolor=blue, urlcolor=blue, pagebackref=true]{hyperref}
\definecolor[named]{ACMDarkBlue}{cmyk}{1,0.58,0,0.21}
\definecolor{darkgreen}{RGB}{25,200,25}
\hypersetup{colorlinks,
  linkcolor=ACMDarkBlue,  
  citecolor=ACMDarkBlue,  
  urlcolor=ACMDarkBlue,   
  filecolor=ACMDarkBlue
}

\usepackage{cleveref}
\Crefformat{figure}{Fig.~#2#1#3}
\Crefformat{equation}{Eq.~#2#1#3}
\Crefformat{table}{Tab.~#2#1#3}

\usepackage{caption}
\usepackage{tabularx,booktabs}
\newcolumntype{Y}{>{\centering\arraybackslash}X}
\usepackage{multirow}
\usepackage{soul}
\newcommand{\drule}{\specialrule{0.2pt}{1pt}{1pt}%
            \specialrule{0.2pt}{0pt}{\belowrulesep}%
            }

\usepackage{algorithm,algpseudocode}
\usepackage[algo2e]{algorithm2e} 
\definecolor{commentcolor}{RGB}{110,154,155}
\definecolor{defcolor}{RGB}{225,81,145}

\let\titleold\title
\renewcommand{\title}[1]{\titleold{#1}\newcommand{\thetitle}{#1}}
\def\maketitlesupplementary
   {
       \twocolumn[
        \centering
        \Large
        \textbf{\thetitle}\\
        \vspace{0.5em}Supplementary Material \\
        \vspace{1.0em}
       ] 
   }

\usepackage{pifont}
\renewcommand{\cmark}{\textcolor{ForestGreen}{\ding{51}}}%
\renewcommand{\xmark}{\textcolor{BrickRed}{\ding{55}}}%

\definecolor{color_mhred}{RGB}{230,34,46}
\definecolor{color_mhblue}{RGB}{0,90,190}
\definecolor{color_mhgreen}{RGB}{112,173,71}

\usepackage{anyfontsize}
\usepackage{enumitem}

\usepackage{amsmath,amsfonts,bm}

\def\eqref#1{equation~\ref{#1}}

\def\floor#1{\lfloor #1 \rfloor}
\def\1{\bm{1}}

\def\vd{{\bm{d}}}

\def\vf{{\bm{f}}}

\def\vl{{\bm{l}}}

\def\vp{{\bm{p}}}

\def\vs{{\bm{s}}}

\def\vu{{\bm{u}}}
\def\vv{{\bm{v}}}

\def\vy{{\bm{y}}}
\def\vz{{\bm{z}}}

\def\mI{{\bm{I}}}

\def\mP{{\bm{P}}}

\def\mS{{\bm{S}}}

\def\mW{{\bm{W}}}

\DeclareMathAlphabet{\mathsfit}{\encodingdefault}{\sfdefault}{m}{sl}
\SetMathAlphabet{\mathsfit}{bold}{\encodingdefault}{\sfdefault}{bx}{n}

\def\gI{{\mathcal{I}}}

\def\gN{{\mathcal{N}}}

\def\gP{{\mathcal{P}}}

\def\gS{{\mathcal{S}}}
\def\gT{{\mathcal{T}}}

\def\gY{{\mathcal{Y}}}
\def\gZ{{\mathcal{Z}}}

\def\sP{{\mathbb{P}}}

\def\sS{{\mathbb{S}}}

\newcommand{\R}{\mathbb{R}}

\newcommand\norm[1]{\left\lVert#1\right\rVert}
\def\degree{{{}^{\circ}}}

\usepackage{nameref}

\title{
SphereDiff: Tuning-free 360$\degree$ Static and Dynamic Panorama Generation \\
via Spherical Latent Representation
}
\author{
    Minho Park$^*$,
    Taewoong Kang$^*$,
    Jooyeol Yun,
    Sungwon Hwang,
    Jaegul Choo
}
\affiliations{
    KAIST AI, South Korea\\
    \{m.park, keh0t0, blizzard072, shwang.14, jchoo\}@kaist.ac.kr
}

\usepackage{graphicx}
\usepackage{animate}
\usepackage{cuted}
\usepackage{tabularx}
\usepackage{capt-of}

\usepackage{balance}

\begin{document}

\twocolumn[{%
\renewcommand\twocolumn[1][]{#1}%
\maketitle
\begin{center}
\centering
\captionsetup{type=figure}
\includegraphics[width=\linewidth]{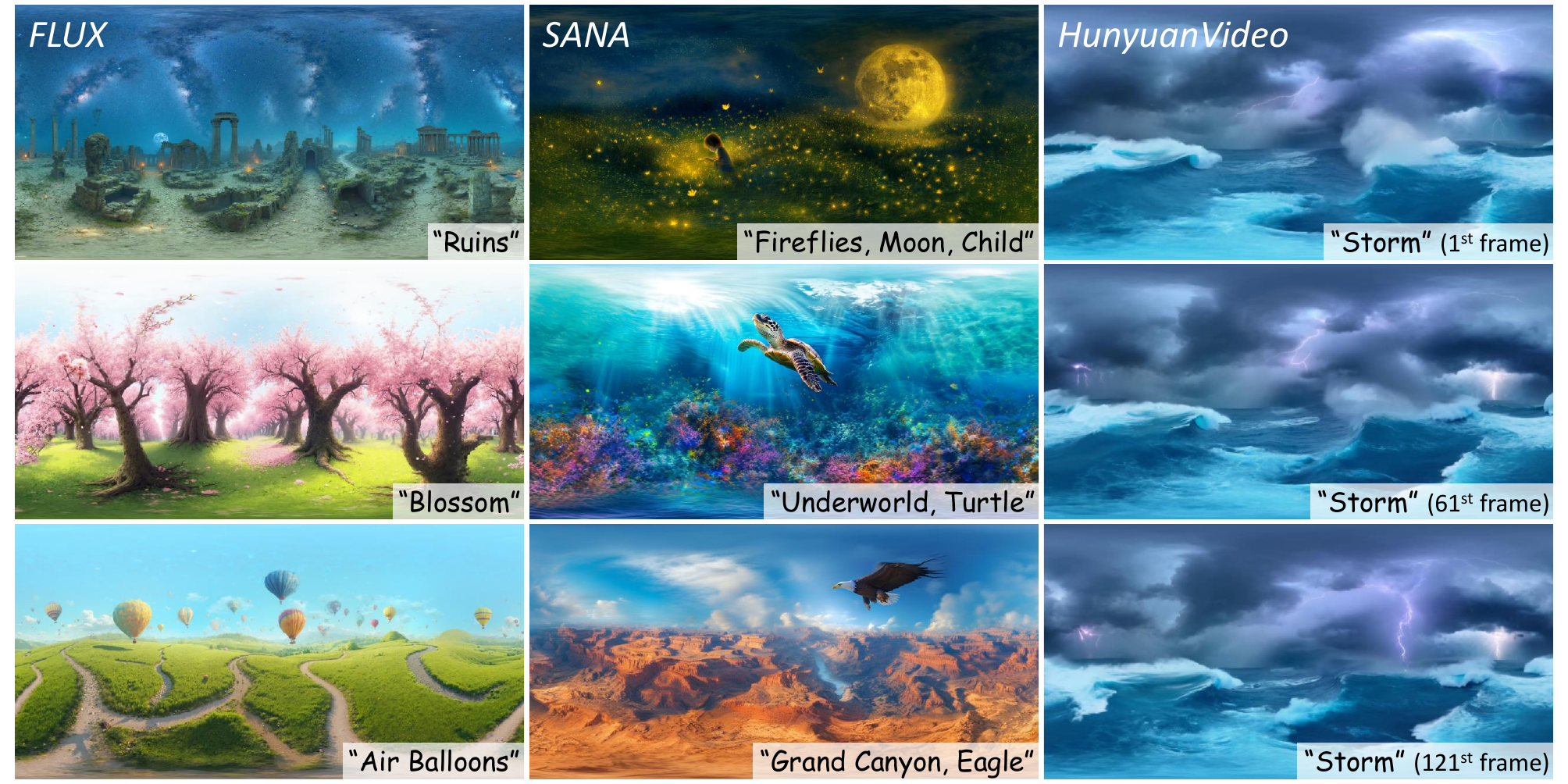}
\vspace{-0.4cm}
\captionof{figure}{
\textbf{SphereDiff enables tuning-free 360° panorama generation via spherical latent.}
It is compatible with various diffusion backbones, including FLUX~\cite{flux2024}, SANA~\cite{xie2024sana}, and HunyuanVideo~\cite{kong2024hunyuanvideo}.
}
\label{fig:1_teaser}
\vspace{0.1cm}
\end{center}
}]

\begin{abstract}
The increasing demand for AR/VR applications has highlighted the need for high-quality content, such as 360$\degree$ live wallpapers. 
However, generating high-quality 360$\degree$ panoramic contents remains a challenging task due to the severe distortions introduced by equirectangular projection (ERP).
Existing approaches either fine-tune pretrained diffusion models on limited ERP datasets or adopt tuning-free methods that still rely on ERP latent representations, often resulting in distracting distortions near the poles.
In this paper, we introduce \textit{SphereDiff}, a novel approach for synthesizing 360$\degree$ static and live wallpaper with state-of-the-art diffusion models without additional tuning. 
We define a spherical latent representation that ensures consistent quality across all perspectives, including near the poles.
Then, we extend MultiDiffusion to spherical latent representation and propose a dynamic spherical latent sampling method to enable direct use of pretrained diffusion models.
Moreover, we introduce distortion-aware weighted averaging to further improve the generation quality.
Our method outperforms existing approaches in generating 360$\degree$ static and live wallpaper, making it a robust solution for immersive AR/VR applications.
\footnote{
\textbf{Links:}
\href{https://github.com/pmh9960/SphereDiff}{Github Code} $\mid$ 
\href{https://pmh9960.github.io/research/SphereDiff/}{Project Page}\\
\makebox[0.5cm][r]{}* indicates equal contributions.
}
\end{abstract}

\vspace{-0.3cm}
\section{Introduction}
\label{sec:introduction}

The growing demand for AR/VR applications has significantly increased the need for high-quality immersive content. 
AR/VR technologies offer highly engaging environments, providing a sense of presence that traditional displays (\eg, phones and laptops) cannot.
A key element in delivering such experiences is the $360^{\circ} \times 180^{\circ}$ panoramic scene, or 360$\degree$ panorama, which provides an omnidirectional view of the virtual world.
This allows users to explore their surroundings from any perspective, setting it apart from standard visual content.
However, because capturing 360$\degree$ panoramas requires specialized cameras, their availability is limited, especially for videos.
As a result, available content is dominated by simulation-based graphics, which can be rudimentary, while users increasingly seek realistic experiences.
In this paper, we aim to generate realistic 360$\degree$ static and live wallpapers (\Cref{fig:1_teaser}), enabling the creation of countless immersive scenes without specialized cameras.

360$\degree$ panoramas are typically represented using an equirectangular projection (ERP), which maps spherical imagery onto a 2D rectangular plane, \eg, mapping a 3D globe to a 2D world map.
Due to the limited representational capacity of a 2D plane, an ERP inevitably introduces severe nonlinear distortions, known as ERP distortion, where high-latitude regions appear disproportionately large.
For example, as shown in \Cref{fig:1_teaser}, the content near the poles appear significantly larger than the others since we visualize the 360$\degree$ wallpapers in ERP.
Due to this ERP distortion, 360$\degree$ panoramas lie in a significantly different distribution from standard perspective images or videos, making it challenging to leverage standard pretrained image or video diffusion models~\cite{ldm, hacohen2024ltx}.

\begin{figure}[t]
\centering
\includegraphics[width=\linewidth]{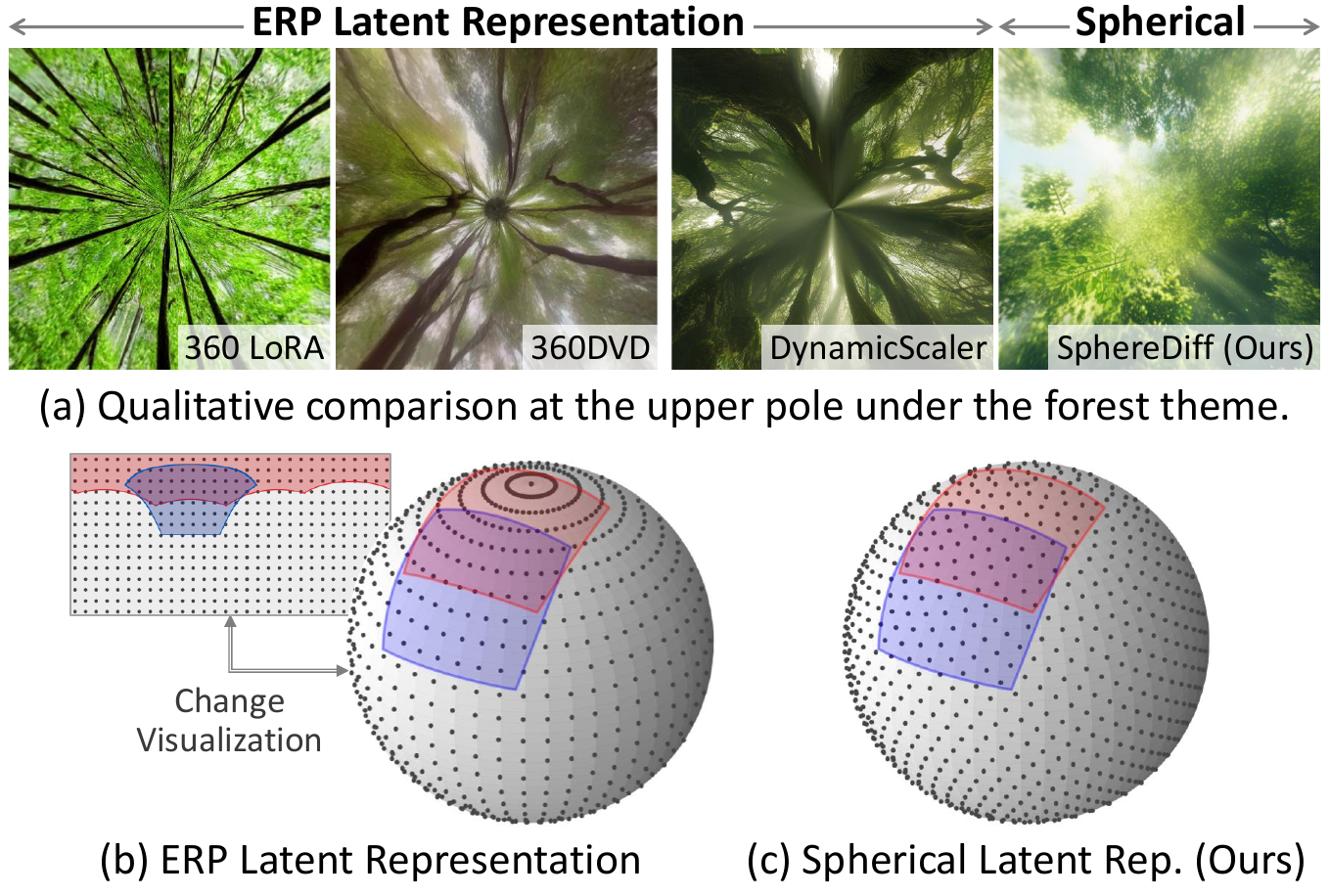}
\vspace{-0.4cm}
\caption{
\textbf{Motivation.}
Both ERP-based finetuning~\cite{360_lora,wang2024360dvd} and tuning-free~\cite{liu2024dynamicscaler} approaches often fail to generate seamless scenes near the poles, as their latents are unevenly distributed over the spherical surface.
In contrast, our method produces seamless results by leveraging a spherical latent representation.
}
\vspace{-0.25cm}
\label{fig:1_motivation}
\end{figure}

To handle this gap, several previous studies have fine-tuned pretrained diffusion models using ERP datasets~\cite{360_lora, wang2024360dvd, chen2022text2light, zhang2024panfusion, li20244k4dgen}.
However, due to the limited availability of text-ERP pairs, data-driven approaches often fail to generate seamless 360$\degree$ panoramas, particularly near the poles, as shown in \Cref{fig:1_motivation}.
Notably, this issue is more pronounced for generating dynamic 360$\degree$ panoramas due to the \textit{severely limited availability of 360$\degree$ panoramic videos.}
Thus, tuning-free approaches offer an alternative approach for generating 360$\degree$ live wallpapers.

Previous tuning-free approaches are based on MultiDiffusion~\cite{liu2024dynamicscaler, bar2023multidiffusion}, which denoises large panoramic latents by dividing them into small overlapping patches and blending them in the overlapping regions.
They use an ERP representation for latents, that distributes latents uniformly over the ERP.
Despite its simplicity, it causes significant differs the density of latents in spherical representation, resulting severe pole-stretching artifacts, as shown in \Cref{fig:1_motivation}.
In this paper, we present a novel tuning-free framework, \textit{SphereDiff}, that does not rely on the ERP representation, and generates seamless 360$\degree$ static and live wallpapers with minimal distortion, even near the poles.

To this end, we define a spherical latent representation that \textit{uniformly distributes latents over the sphere}, ensuring consistent generation quality across all view directions.
We then extend the tuning-free MultiDiffusion framework~\cite{bar2023multidiffusion} to operate within this spherical latent representation.
In addition, we propose a dynamic latent sampling algorithm that effectively arranges spherical latents onto a 2D perspective grid.
Finally, we introduce a distortion-aware weighted averaging scheme to further reduce minor distortions caused by spherical-to-perspective projection.
Extensive experiments demonstrate that SphereDiff outperforms existing methods in generating static and live 360$\degree$ wallpapers, in terms of visual quality and distortion reduction.

In summary, our contributions are threefold:
\begin{itemize}
    \item We propose \textit{SphereDiff}, a tuning-free framework with a \textit{spherical latent representation} for generating high-quality 360$\degree$ wallpapers, especially near the poles.
    \item We extend MultiDiffusion for 360$\degree$ panoramas with \textit{dynamic latent sampling} for seamless integration with standard diffusion models, allowing tuning-free generation.
    \item \textit{Distortion-aware weighted averaging} further mitigates the minor distortion from spherical-to-perspective projection, and significantly enhance visual quality.
\end{itemize}

\begin{figure*}[t]
\centering
\includegraphics[width=0.95\linewidth]{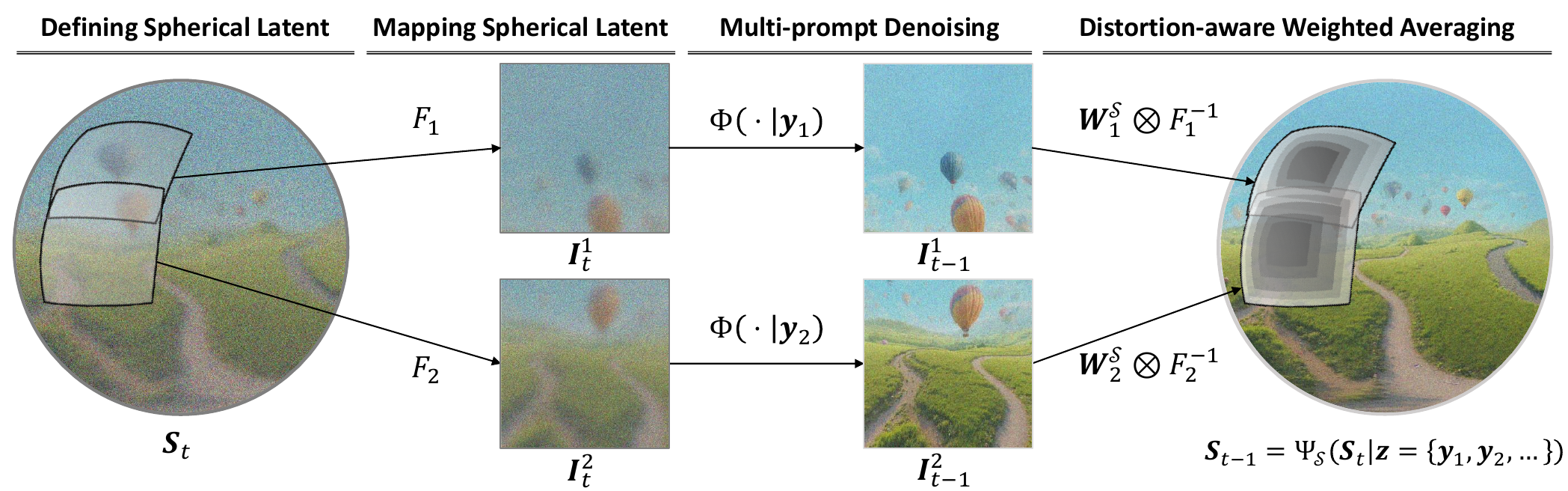}
\caption{\textbf{Overall Pipeline.} 
We begin by initializing uniformly distributed spherical latents.
Next, we map these latents to perspective latents corresponding to multiple view directions.
Each view is then denoised using its corresponding prompt.
The denoised views are subsequently fused via distortion-aware weighted averaging.
}
\label{fig:1_method_overview}
\end{figure*}

\section{Related Work}
\label{sec:related_work}

\paragraph{Latent Diffusion Models.}
Recent advancements in diffusion models have enabled the generation of high-quality images~\cite{ldm, flux2024, xie2024sana, chen2023pixart} and videos~\cite{hacohen2024ltx, yang2024cogvideox, kong2024hunyuanvideo, genmo2024mochi, zhang2025flashvideo, liu2025lumina}, achieving impressive visual results across various video generation tasks within the standard perspective of visual content. However, generating content beyond the standard perspective, such as regular or 360$\degree$ panoramas, remains relatively underexplored.
In this paper, we aim to generate 360$\degree$ panoramas, which differ significantly from standard perspective scenes, by solely leveraging pretrained diffusion models designed for standard perspectives.

\paragraph{360$\degree$ Panoramic Scene Generation.}
Most panoramic generation methods rely on equirectangular projection (ERP), which maps spherical coordinates onto a 2D rectangular plane, with latitude and longitude as the vertical and horizontal axes.
However, ERP inherently introduces severe nonlinear distortions, particularly near the poles, resulting in a significant gap between ERP and standard perspective views.
Although previous studies~\cite{wang2024360dvd, zhang2024panfusion, chen2022text2light, li20244k4dgen} attempt to address this issue by fine-tuning on panoramic ERP datasets, they often fail to generate seamless panoramas, especially near the poles, or struggle with text controllability due to the domain-specific nature of the datasets (\eg, indoor environments).
Recently, CubeDiff~\cite{kalischek2025cubediff} introduces an alternative approach using cube map representations for panoramic image generation.
While this method effectively reduces distortions near the poles by training on the large-scale 360$\degree$ panoramic images, it still struggles with discontinuities at cube-face boundaries.
In addition, the data-driven approaches often struggle on generating 360$\degree$ video, due to the extreme data scarcity of 360$\degree$ panoramic videos.
In contrast, we replace the ERP latent representation with a \emph{spherical latent representation}, providing a natural solution to eliminate distortion across all perspectives, and it does not require any additional tuning.

\paragraph{360$\degree$ Live Wallpaper Generation.}
Due to the limited availability of 360$\degree$ video datasets, recent research on 360$\degree$ live wallpapers has increasingly favored using perspective-based models without additional training.
DynamicScaler~\cite{liu2024dynamicscaler} attempts to mitigate ERP’s inherent distortions through panoramic-projected denoising, leveraging the MultiDiffusion framework~\cite{bar2023multidiffusion} with adjusted windows.
4K4DGen~\cite{li20244k4dgen} seeks to avoid distortion by utilizing the input ERP images with an image-to-video model, which makes it less suitable for generating novel or highly creative content directly from text descriptions.
Unlike these methods, we overcome these limitations by directly using uniformly distributed spherical latents, thereby ensuring efficiency and reduced distortion without requiring additional training.

\section{Proposed Method}
\label{sec:3_method}

In this section, we introduce \textit{SphereDiff}, a novel tuning-free framework for generating 360$\degree$ live wallpapers (\Cref{fig:1_method_overview}).
First, we present the spherical latent representation and spherical-to-perspective projection (\Cref{sec:3_method_spherical_latent}).
Next, we extend the MultiDiffusion framework~\cite{bar2023multidiffusion} for 360$\degree$ panoramas (\Cref{sec:3_method_multi_diffusion}).
We then introduce spherical latent sampling methods, which discrete the continuous coordinates of the spherical latent onto a 2D grid (\Cref{sec:3_method_sampling}).
Finally, we propose a distortion-aware weighted averaging method to mitigate minor distortions from the spherical-to-perspective projection (\Cref{sec:3_method_weighted_averaging}), and introduce the multi-prompt inference method (\Cref{sec:3_method_multi_prompt}).

\subsection{Spherical Latent Representation}
\label{sec:3_method_spherical_latent}

\paragraph{Definition.}
We introduce a spherical representation of latent features for generating 360$\degree$ live wallpapers.
We define a latent feature $\vf \in \mathbb{R}^C$ paired with the corresponding spherical coordinate $\vd$ on a spherical surface.
The set of spherical coordinates can be represented as follows:
\begin{equation}
\mathbb{S}^2 = \{\vd = (x, y, z) \mid x, y, z \in \mathbb{R}, \|\vd\| = 1 \}.
\end{equation}
Then, we pair each latent feature with its associated position, \ie, $\vs = (\vd, \vf)$, referred to as \textit{spherical latent}.
For $N$ spherical latents, we define the spherical latents $\mS$ as:
{\small
\begin{equation}
\mS = \{\vs_i = (\vd_i, \vf_i) \mid \vd_i \in \mathbb{S}^2, \vf_i \in \mathbb{R}^C, \text{ for } i \in [1, N] \}.
\end{equation}}
which is now composed of multiple latents similar to standard 2D or 3D latent features.
We refer to the domain of spherical latents as $\gS$, \ie, $\mS \in \gS$.

Equirectangular Projection (ERP) latents also can be written in our spherical latent representation.
However, as shown in \Cref{fig:1_motivation} (b), due to the 2D grid constraint of ERP latent, its spherical coordinates are not uniformly distributed on the sphere's surface.
In contrast, we define the spherical latents using the Fibonacci Lattice~\cite{fibonacci}, which offers the number of spherical latents is nearly equal across all perspectives, as shown in \Cref{fig:1_motivation} (c).

\paragraph{Perspective Latent Representation.}
Since standard diffusion models operate in perspective space, we utilize a \textit{spherical-to-perspective projection} which transforms the spherical coordinate to the perspective coordinate.
To achieve this, we first define the domain of perspective coordinates as a discretized 2D plane as
{\small
\begin{equation}
\sP^2 = \left\{\vu = \left(\frac{2j}{H}, \frac{2k}{W}\right ) \mid j\in \left[-\frac{H}{2}, \frac{H}{2} \right ], k\in \left [ -\frac{W}{2},\frac{W}{2} \right ] \right\},
\end{equation}}
where $H, W$ indicates the height and width of the bounded 2D perspective plane, respectively.
We use a view direction $\vv \in \mathbb{S}^2$ and a predefined focal length $f$ to define the spherical-to-perspective projection function $\vu = \mathcal{T}_{\mathbb{S}^2 \rightarrow \mathbb{P}^2} (\vd| \vv, f)$.
For completeness, the formula of the projection function is provided in \Cref{appn:spherical_to_perspective}.

\begin{figure}[t]
\centering
\includegraphics[width=0.95\linewidth]{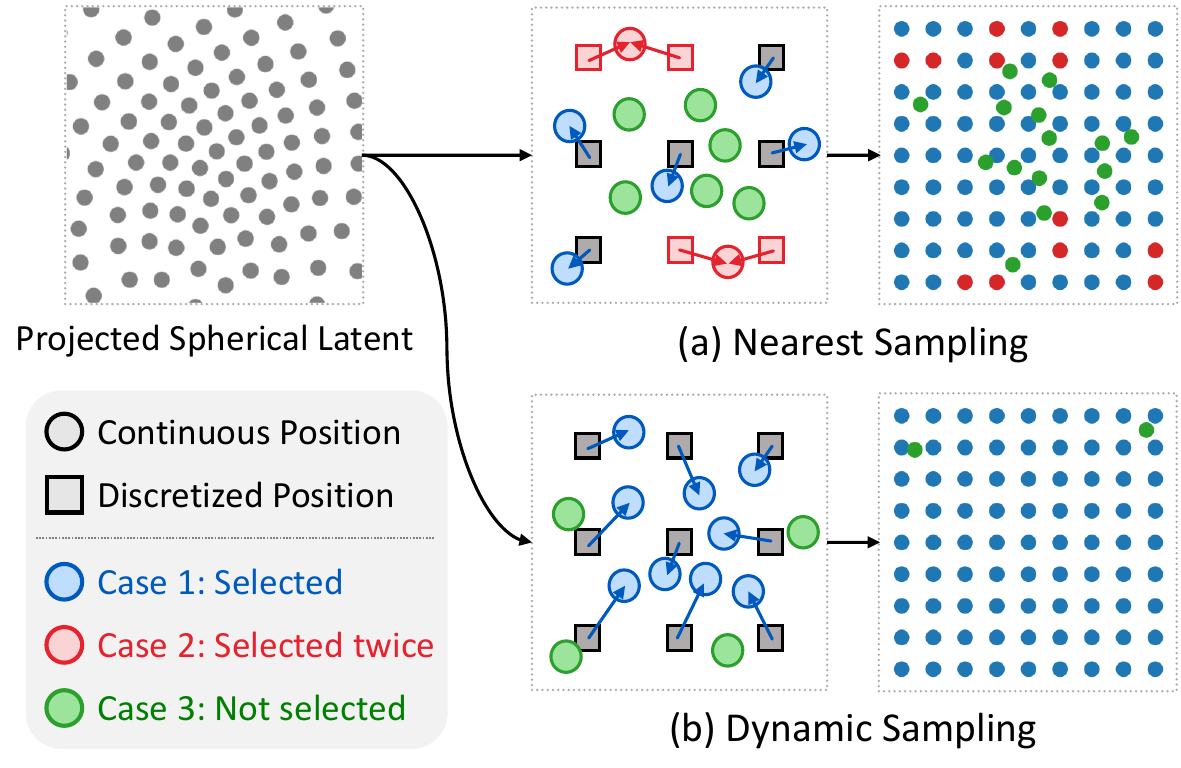}
\caption{
\textbf{Comparison of Nearest and Dynamic Sampling.} Nearest sampling often resamples the selected latents or omits central ones, while dynamic sampling selects latents from the center outward, discarding only the outermost ones.
}
\label{fig:3_discretization}
\end{figure}

\subsection{MultiDiffusion for Spherical Latent}
\label{sec:3_method_multi_diffusion}

The MultiDiffusion~\cite{bar2023multidiffusion} framework is often utilized for generating arbitrary-shaped images by leveraging pretrained diffusion models~\cite{ldm} trained on standard perspective images.
In this section, we introduce an extension of the MultiDiffusion framework to the spherical latent representation.

The goal of this framework is to construct the \textit{Spherical MultiDiffuser} $\Psi_\mathcal{S}: \gS\times \gZ \rightarrow \gS$, which takes a noisy spherical latent $\mS_t$ and a set of text conditions $\vz$ as inputs and produces the denoised spherical latent $\mS_{t-1}$, as illustrated in \Cref{fig:1_method_overview}.
Based on the MultiDiffuser, a clean spherical latent $\mS_0$ can be obtained from pure noise $\mS_T$ through an iterative denoising process using diffusion models as:
\begin{equation}
    \mS_T, \mS_{T-1}, ..., \mS_0 \quad \text{ s.t. } \quad \mS_{t-1} = \Psi_\mathcal{S}(\mS_t | \vz).
\end{equation}

To construct MultiDiffuser $\Psi_\mathcal{S}$, we first leverage a pretrained diffusion model trained on standard perspective latents $\Phi: \gI\times \gY \rightarrow \gI$, which takes a noisy latent $\mI_t$ and a text condition $\vy$ as inputs and produces the denoised latent $\mI_{t-1}$.
The pretrained diffusion model gradually denoises the pure Gaussian noise $\mI_T\sim \gN$ into a clean image $\mI_0$.
\begin{equation}
    \mI_T, \mI_{T-1}, ..., \mI_0 \quad \text{ s.t. } \quad \mI_{t-1} = \Phi(\mI_t | \vy)
\end{equation}

Next, we define a mapping function between the spherical and perspective latent spaces, $F_i: \gS \rightarrow \gI$, along with a corresponding condition mapping $\lambda_i: \gZ \rightarrow \gY$, where $i \in \{1, ..., n\}$.
The mapping functions $F_i$ and $\lambda_i$ can be formulated in various ways, which will be discussed in \Cref{sec:3_method_sampling,sec:3_method_multi_prompt}, respectively.
\begin{equation}
    \mI_t^i = F_i (\mS_t), \quad \vy_i = \lambda_i (\vz)
\end{equation}

Finally, The denoising step in of MultiDiffuser can be formulated by a closed-form~\cite{bar2023multidiffusion}.
\begin{equation}
    \Psi(\mS_t | \vz) = \sum_{i=1}^n \mW^{\gS}_i \otimes F_i^{-1} (\Phi(\mI_t^i | \vy_i)).
\end{equation}
where $\mW^{\gS}_i$ are the per-pixel weights and $\otimes$ is the Hadamard product.
In the following sections, we define the latent mapping function $F_i$ (\Cref{sec:3_method_sampling}), the per-pixel weights $\mW^{\gS}_i$ (\Cref{sec:3_method_weighted_averaging}), and the condition mapping function $\lambda_i$ (\Cref{sec:3_method_multi_prompt}), which together extend MultiDiffusion to the spherical latent representation.

\begin{algorithm}[t]
\SetKwInOut{KwIn}{Input}
\SetKwInOut{KwOut}{Output}
\SetKwFunction{Sort}{SortByOrigin}
\SetKwData{H}{H}
\SetKwData{W}{W}
\caption{Dynamic Latent Sampling}
\label{algo:dynamic_sampling}

\KwIn {Projected Latent $\mP = \gT_{\gS\rightarrow\gI}(\mS|\vv, f)$}
\KwOut {Arranged Perspective Latent $\mI$}

$\mI' \longleftarrow$ Sort the latents of $\mP$ by $\norm{\vu_i}$.

$M \longleftarrow |\mP|$ \Comment{Get the number of latents}

$H, W \longleftarrow \floor{\sqrt{M}}, \floor{\sqrt{M}}$ \Comment{Dynamic $H, W$}

$\mI \longleftarrow \emptyset^{H\times W}$ \Comment{Initialize a queue}

\For{$i \in [1, H / 2]$}{
    $n \longleftarrow (2i)^2 - (2i-2)^2$ \Comment{Counts of $i$-th border}

    $\vl \longleftarrow$ first $n$ latents from the sorted $\mI'$ \Comment{center-first}

    Set $i$-th border of $\mI$ to $\vl$

    Pop first $n$ latents from $\mI'$
}

\Return $\mI$ \Comment{Ignore $M - H\times W$ elements of $\mI$}
\end{algorithm}

\subsection{Mapping Spherical Latent}
\label{sec:3_method_sampling}
We define the latent mapping function $F$ that transforms a spherical latent representation into a perspective latent space $\gI$.
To define mapping function $F$, we first apply the transformation $\gT_{\gS\rightarrow\gI}$ based on the view direction $\vv\in \sS^2$ and focal length $f$, which projects the coordinates of the spherical latents onto the perspective plane $\gP$.
Formally, the spherical-to-perspective latent transformation can be written as
{\small
\begin{equation}
\gT_{\gS\rightarrow\gP}(\mS|\vv, f) = \mP = \{ \vp_i = (\vu_i, \vf_i) | \vu_i \in [-1, 1]^2 \},
\end{equation}}
where $\vu_i = \gT_{\sS^2 \rightarrow \sP^2}(\vd_i|\vv, f)$.
Note that, $\mP$ does not contain $N$ elements since the perspective are cropped by $[-1, 1]$.
We next introduce simple yet effective latent sampling methods for discretizing perspective coordinates.

\begin{figure*}[t]
\centering
\includegraphics[width=0.95\linewidth]{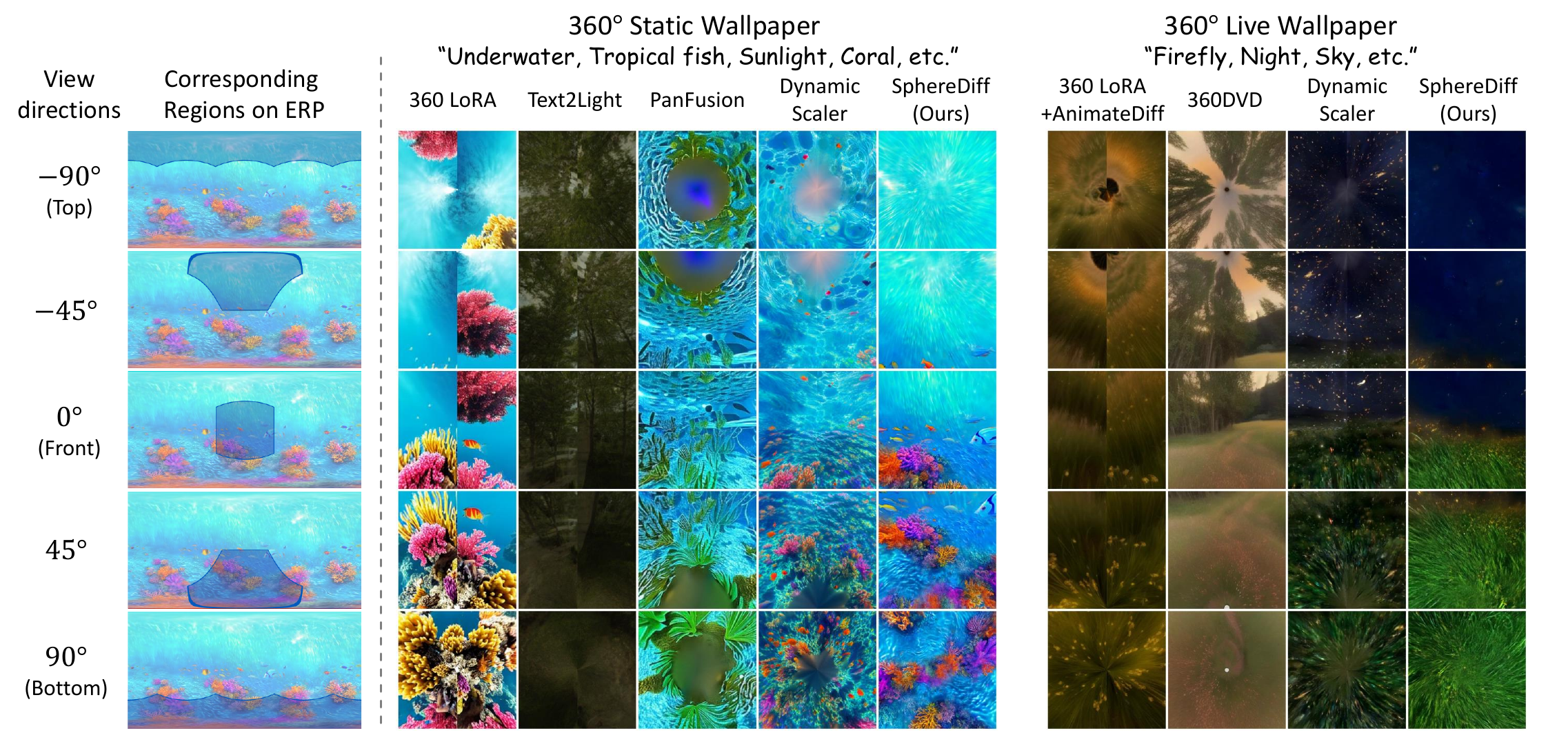}
\caption{
\textbf{Qualitative comparison.}
Each sample shows perspective views from top to bottom, highlighting end-to-end continuity and distortion.
Other methods exhibit artifacts such as seams, pole distortions, blurriness, or spots, while ours produces seamless, high-quality panoramas without these issues.
The entire ERPs are available in \Cref{appn:comparison}.
}
\label{fig:4_quali_comparison}
\end{figure*}

\paragraph{Nearest Point Sampling.}
A straightforward approach to discrete continuous coordinates is nearest-neighbor sampling, where the nearest projected spherical latent is selected for each pixel position.
Specifically, the latent closest to the center of a $H \times W$ grid is retrieved and used as input for denoising, as illustrated in \Cref{fig:3_discretization} (a).
Despite its simplicity, this method introduces two critical issues. First, the same latent may be selected multiple times, altering the latent distribution, which often degrades generation performance~\cite{chang2024warped}.
Second, some spherical latents may not be chosen even if they fall within the field of view of the current camera view direction.
This phenomenon, referred to as the \emph{undersampling problem}, has particularly detrimental consequences for generating a seamless panorama.

\paragraph{Undersampling Problem.}
The undersampling of spherical latents disrupts information flow across neighboring windows.
As illustrated in \Cref{fig:3_discretization}, the green points lack information from the current field of view (FoV) since they are not denoised in this step.
If the next window's FoV captures a green point, not the blue point, it receives no information from the current window, \emph{causing discontinuities even when there is a large overlap.}

\paragraph{Dynamic Latent Sampling.}
To address the undersampling problem, we aim to ensure that all points within the FoV are selected, especially those near the center.
We propose a dynamic latent sampling strategy, which comprises three components: (1) a queue, (2) a dynamic number of latents, and (3) center-first selection.
First, we avoid selecting the same spherical latent more than once by using a queue: once a latent is selected, it is immediately removed from the queue.
Then, we dynamically adjust the number of latents so that $H$ and $W$ are not fixed, thereby reducing the number of latents within the FoV that remain unselected.
Lastly, we prioritize selecting center-positioned latents first and then ignore the remaining points at the outermost region.
The entire algorithm is demonstrated in \Cref{algo:dynamic_sampling} and \Cref{fig:3_discretization} (b).

\subsection{Distortion-Aware Weighted Averaging}
\label{sec:3_method_weighted_averaging}

While the spherical-to-perspective distortion is relatively smaller than the ERP-to-perspective distortion, it can still cause latent position misalignment for other viewpoints.

To address this, we proposed distortion-aware weighted averaging within the MultiDiffusion framework~\cite{bar2023multidiffusion}.
Specifically, we adjust the per-pixel weight $\mW^\gS_i$ to account for the spherical-to-perspective distortion.
Since distortion increases with distance from the center of the perspective image, we introduce a simple yet effective exponential weighting function in the image space $\gI$.
\begin{align}
    \mW^\gI_i &= [W^{jk}_i]_{j\in[1,H],k\in[1,W]} \in \R^{H\times W}, \\
    W^{jk}_i &= \exp \left( - \norm{\vu_{jk}} / \tau \right),
\end{align}
where $\norm{\vu_{jk}}$ is the distance from the center of the perspective image, and $\tau$ is a scaling factor controlling how quickly the weight decays toward the edges.
Then, the weighting function can be represented as $\mW^\gS_i = F^{-1}_i(\mW^\gI_i)$.

\begin{table*}[t]
\centering
\resizebox{0.92\linewidth}{!}{
\begin{tabular}{c|l|cccccc}

\toprule
& & \multicolumn{2}{c}{Panoramic Criteria} & Image Criteria & Text Adherence & \multicolumn{2}{c}{Video Criteria} \\
\cmidrule(lr){3-4} \cmidrule(lr){5-5} \cmidrule{6-6} \cmidrule(lr){7-8}
Wallpaper Type & Method & Distortion $\uparrow$ & End Continuity $\uparrow$ & Image Quality $\uparrow$ & Text Alignment $\uparrow$ & \begin{tabular}[c]{@{}c@{}}Motion\\ Smoothness\end{tabular} $\uparrow$ & \begin{tabular}[c]{@{}c@{}}Temporal\\ Flickering\end{tabular} $\uparrow$ \\
\drule
& 360 LoRA & \underline{21.43} & 23.81 & 21.43 & 20.24 & - & - \\
& Text2Light & 5.95 & 4.76 & 8.33 & 5.95 & - & - \\
360$\degree$ Static & PanFusion & 14.29 & 10.71 & 10.71 & 16.67 & - & - \\
& DynamicScaler & 20.24 & \underline{25.00} & \underline{25.00} & \underline{27.38} & - & - \\
& SphereDiff (Ours) & \textbf{38.10} & \textbf{35.71} & \textbf{34.52} & \textbf{29.76} & - & - \\
\drule
\multirow{4}{*}{360$\degree$ Live} & 360 LoRA + AnimateDiff & 25.00 & \underline{27.38} & \underline{27.38} & \underline{27.38} & 27.38 & 23.81 \\
& 360DVD & 11.90 & 13.10 & 16.67 & 20.24 & 8.33 & 14.29 \\
& DynamicScaler & \underline{30.95} & 26.19 & \textbf{28.57} & 22.62 & \underline{28.57} & \underline{29.76} \\
& SphereDiff (Ours) & \textbf{32.14} & \textbf{33.33} & \underline{27.38} & \textbf{29.76} & \textbf{35.71} & \textbf{32.14} \\
\bottomrule

\end{tabular}
}
\caption{
\textbf{User study results.}
The 360$\degree$ static and live wallpapers generated by SphereDiff have achieved state-of-the-art performance in user preference across most metrics, particularly in panoramic criteria such as distortion and end continuity.
}
\label{tab:exp_user_study}
\end{table*}

\subsection{Multi-prompt Inference}
\label{sec:3_method_multi_prompt}

We generate 360$\degree$ wallpapers using multiple prompts that correspond to specific regions on the spherical surface.
The use of region-specific prompts helps reduce semantic inconsistencies, such as generating ground in sky regions.
To this end, we define a simple condition mapping function $\lambda_i$ that selects a region-specific prompt from the prompt set $\vz$, according to the elevation of the view direction, $\phi_i = \operatorname{elevation}(\vd_i)$.
Specifically, $\lambda_i$ selects the prompt whose elevation is closest to $\phi_i$ among the discrete elevations $\{-90\degree, -10\degree, 0\degree, +10\degree, +90\degree\}$, with the corresponding prompts given by $\vz = \{\vy_\text{top}, \vy_\text{upper}, \vy_\text{middle}, \vy_\text{lower}, \vy_\text{bottom}\}$.
We allocate denser text prompts near the horizon to capture richer visual complexity in that region.

For enhanced scene complexity, we employ an additional foreground prompt $\vy_\text{foreground}$, conditioned on the corresponding view direction, to generate a foreground object at a specific location (\eg, $\theta_i = \operatorname{azimuth}(\vd_i) = 30\degree, \phi_i = -30\degree$).
The second column of \Cref{fig:1_teaser} shows an example generated in this manner.
Visual illustrations of the multi-prompt inference are provided in \Cref{appn:multi_prompt}.

\section{Experiments}
\label{sec:experiments}

\subsection{Experimental Setup}

\paragraph{Implementation Details.}

For all experiments and comparisons, we adopt SANA~\cite{xie2024sana} and LTX-Video~\cite{hacohen2024ltx} as the base T2I and T2V models, respectively.
We use 2,600 spherical latent points, and we conduct the MultiDiffusion framework with 89 view directions with an 80$\degree$ FoV, where adjacent views overlap by 60\%.
The additional details are available in \Cref{appn:implementation_details}.

\paragraph{Evaluation Criteria.}

We evaluate our 360$\degree$ static and live wallpapers using four criteria: panoramic, image, video, and text-level aspects.
Image quality, text adherence, and temporal smoothness are standard metrics in image and video generation, and we adopt them to assess the realism and consistency of our results.
For panoramic evaluation, we consider distortion and end continuity.
Distortion measures the geometric deformation introduced when the equirectangular projections (ERPs) are converted into perspective, which tends to increase if models fail to account for ERP-specific constraints.
End continuity (also known as loop-consistency) evaluates the seamless alignment of the left and right borders in ERPs, indicating whether the scene wraps smoothly into a loop.

\paragraph{Evaluation Process.}

We use 20 predefined text prompt sets designed for immersive outdoor scenes.
We assess the criteria through a user study, following prior studies~\cite{wang2024360dvd, liu2024dynamicscaler}, where participants select the sample among the baselines that best fits the given criteria.
The assessments are conducted on perspective images captured from 14 predefined view directions with a 90$\degree$ field of view.
Additionally, we conduct automatic evaluations for text and video criteria by leveraging VBench~\cite{huang2024vbench}, which has been widely validated.
In contrast, image and panoramic criteria cannot be accurately measured by classical methods such as FID~\cite{heusel2017gans} due to the absence of a source domain.
Thus, we instead employ vision-language models~\cite{gpt4o} within the LLM-as-a-judge framework~\cite{zheng2023judging}.
The reliability of the VLM-based evaluation, supported by comprehensive experiments, is provided in \Cref{appn:evaluation_details}.

\paragraph{Baselines.}
For 360$\degree$ static wallpaper generation, we compare with open-sourced baselines including 360 LoRA~\cite{360_lora}, Text2Light~\cite{chen2022text2light}, Panfusion~\cite{zhang2024panfusion}, and DynamicScaler~\cite{liu2024dynamicscaler}.
Although DynamicScaler~\cite{liu2024dynamicscaler} was originally designed for panoramic video generation, we extend it for image generation and use it as a baseline.
For 360$\degree$ live wallpaper generation, we use 360DVD~\cite{wang2024360dvd} and DynamicScaler as primary baselines.
DynamicScaler is reimplemented with SANA and LTX-Video to enable a fair comparison as a tuning-free method.
In addition, we combine 360 LoRA with AnimateDiff~\cite{guo2023animatediff} as an extra baseline.
The multi-prompt methods, DynamicScaler and ours, share identical text prompt sets ($\vz = \{\vy_\text{top}, \vy_\text{upper}, \vy_\text{middle}, \vy_\text{lower}, \vy_\text{bottom}\}$) for generation, whereas the other single-prompt methods rely on the main reference prompt ($\vy_\text{middle}$).

\subsection{Results}

\paragraph{Qualitative Comparison.}

As shown in \Cref{fig:4_quali_comparison}, 360$\degree$ static and live wallpaper generation baselines appear noticeable artifacts near the poles, such as distortion, blurriness, and speckling due to the limitation of the ERP latents as we discussed.
These artifacts significantly undermine the immersive experience of 360$\degree$ wallpapers.
In contrast, our method ensures consistent quality at all viewing angles and eliminating distortions and discontinuities, since we utilize uniformly distributed spherical latents.
Additional qualitative results, including samples generated by stronger diffusion backbones, namely FLUX~\cite{flux2024} and HunyuanVideo~\cite{kong2024hunyuanvideo}, are provided in \Cref{appn:comparison}.

\paragraph{User Study.}

To evaluate the effectiveness of our panorama generation method, we conduct a user study.
A total of 21 participants compared 20 pairs of samples, both images and videos, based on six aspects: visual quality, text alignment, distortion, end continuity, motion smoothness, and temporal flickering.
As shown in \Cref{tab:exp_user_study}, our method consistently outperforms all baselines on most metrics for both static and live 360$\degree$ wallpaper generation.
While DynamicScaler~\cite{liu2024dynamicscaler} received slightly higher scores in image quality, our method achieves significantly better performance in panoramic metrics.
Overall, the user study strongly supports the effectiveness of the proposed method, as it demonstrates the best results in panoramic criteria, text adherence and video criteria, even without any additional finetuning.

\begin{table*}[t]
\centering
\resizebox{\linewidth}{!}{
\begin{tabular}{c|l|cccccccc}

\toprule
& & \multicolumn{2}{c}{Panoramic Criteria} & \multicolumn{2}{c}{Image Criteria} & \multicolumn{2}{c}{Text Adherence} & \multicolumn{2}{c}{Video Criteria} \\
\cmidrule(lr){3-4} \cmidrule(lr){5-6} \cmidrule(lr){7-8} \cmidrule(lr){9-10}
Wallpaper Type & Method & Distortion $\uparrow$ & End Continuity $\uparrow$ & \begin{tabular}[c]{@{}c@{}}Image\\ Quality\end{tabular} $\uparrow$ & \begin{tabular}[c]{@{}c@{}}Aesthetic\\ Appearance\end{tabular} $\uparrow$ & Scene $\uparrow$ & CLIP-Score $\uparrow$ & \begin{tabular}[c]{@{}c@{}}Motion\\ Smoothness\end{tabular} $\uparrow$ & \begin{tabular}[c]{@{}c@{}}Temporal\\ Flickering\end{tabular} $\uparrow$ \\
\drule
& 360 LoRA & 2.027 & 3.423 & 2.965 & 3.492 & \underline{0.2875} & 26.40 & - & - \\
& Text2Light & 2.381 & 3.454 & 2.415 & 2.777 & 0.0250 & 20.46 & - & - \\
360$\degree$ Static & PanFusion & 1.965 & 3.696 & 2.819 & 3.450 & 0.2125 & 25.70 & - & - \\
& DynamicScaler & \underline{2.854} & \underline{3.985} & \textbf{4.496} & \underline{4.577} & 0.2750 & \underline{26.63} & - & - \\
& SphereDiff (Ours) & \textbf{3.238} & \textbf{4.892} & \textbf{4.496} & \textbf{4.685} & \textbf{0.5875} & \textbf{28.65} & - & - \\
\drule
\multirow{4}{*}{360$\degree$ Live} & 360 LoRA + AnimateDiff & 1.939 & \underline{3.482} & \textbf{3.179} & \underline{3.571} & 0.2914 & 26.34 & 0.9908 & 0.9847 \\
& 360DVD & \underline{2.086} & 3.246 & 2.929 & 3.396 & 0.2570 & 25.54 & 0.9857 & 0.9798 \\
& DynamicScaler & 1.971 & 2.971 & 2.711 & 3.236 & \underline{0.4836} & \underline{26.89} & \underline{0.9943} & \underline{0.9918} \\
& SphereDiff (Ours) & \textbf{2.579} & \textbf{4.496} & \underline{3.050} & \textbf{3.593} & \textbf{0.5703} & \textbf{27.52} & \textbf{0.9956} & \textbf{0.9941} \\
\bottomrule

\end{tabular}
}
\caption{
\textbf{Automated Quantitative Evaluation.} We conduct automatic evaluations of 360$\degree$ static and live wallpaper synthesis across four criteria. SphereDiff consistently outperforms existing methods except image quality, where it ranks second.
}
\label{tab:exp_quanti}
\end{table*}

\paragraph{Automatic Quantitative Results.}

As presented in \Cref{tab:exp_quanti}, our method outperforms all baselines across most evaluation metrics in both static and live 360$\degree$ wallpaper generation, consistent with the results of the user study.
Notably, it achieves significantly better scores in distortion and end continuity, demonstrating its effectiveness in producing high-quality panoramic content.
While the image quality of our video generation is lower than that of 360 LoRA + AnimateDiff~\cite{360_lora,guo2023animatediff}, it remains comparable overall.
The overall score of 360$\degree$ live wallpaper generation is lower than the 360$\degree$ static wallpaper generation, which may be due to the performance of the underlying diffusion model.
Nevertheless, SphereDiff shows the state-of-the-art performance in the most metrics and its performance could be further improved by leveraging a more advanced denoising model.

\begin{figure}[t]
\centering
\includegraphics[width=0.9\columnwidth]{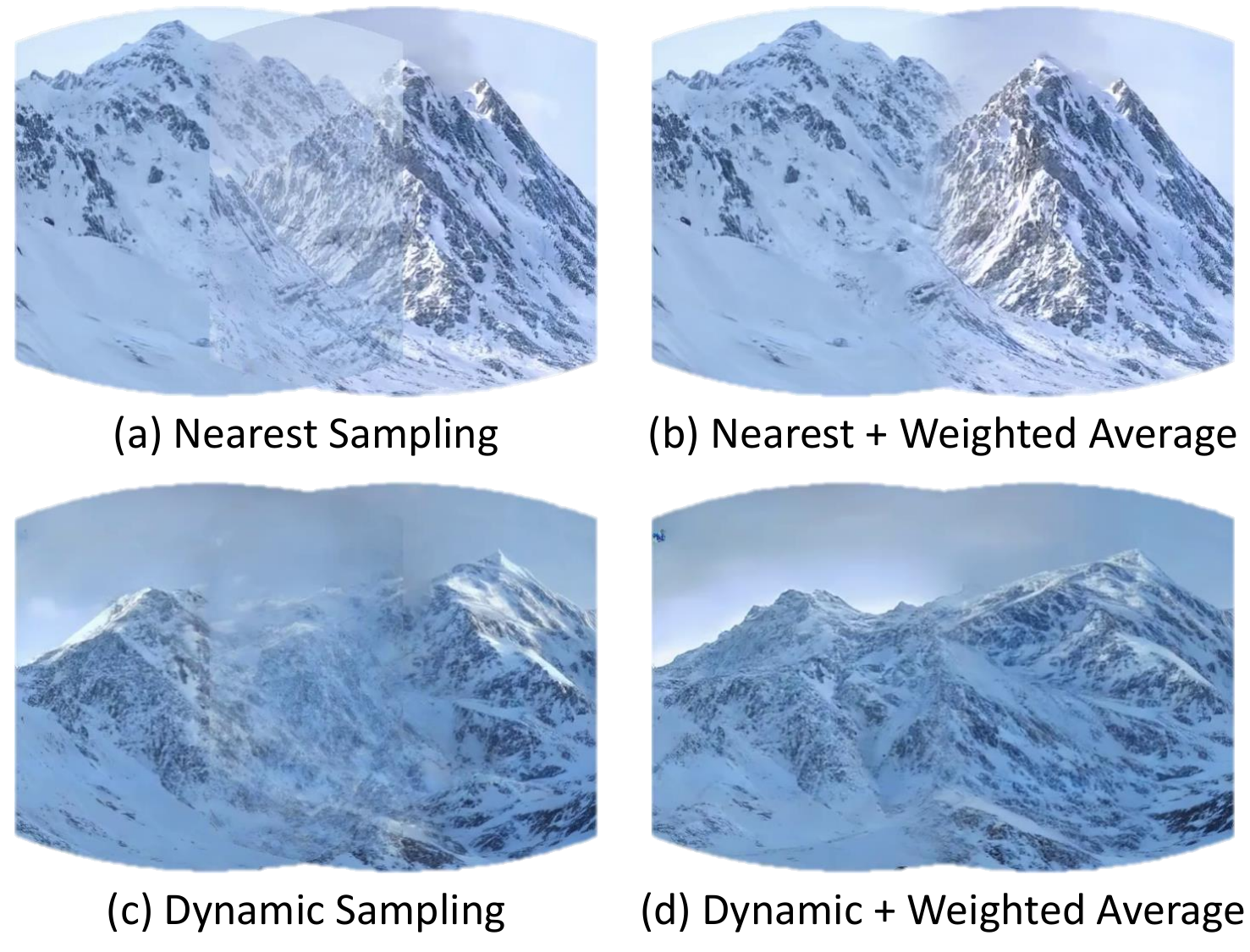}
\caption{
\textbf{Visual Ablation Study.}
Nearest sampling causes view inconsistencies and overlap artifacts due to undersampling, while dynamic sampling enables better information sharing for more integrated outputs.
In addition, weighted averaging significantly improves seamlessness.
}
\vspace{-0.15cm}
\label{fig:4_ablation}
\end{figure}

\subsection{Ablation Study}
\label{sec:exp_ablation}

We conducted ablation studies to evaluate the impact of each component, namely dynamic latent sampling and distortion-aware weighted averaging.
For clarity, we performed denoising on only two views, which simplifies visualization while preserving the validity of the comparison, and we visualized the results in ERP format.
As shown in \Cref{fig:4_ablation}, nearest sampling fails to facilitate information exchange between different views, resulting in noticeable artifacts and unnatural transitions in overlapping regions.
In contrast, our dynamic latent sampling improves information exchange across views, producing more seamless and coherent images.
Furthermore, incorporating our distortion-aware weighted averaging technique yields significantly clearer and more consistent outputs for both sampling methods.
These results demonstrate that each component plays a critical role in our framework by integrating multiple perspectives and ensuring high-quality 360$\degree$ wallpaper generation.
For further evidence, a comprehensive quantitative comparison of the ablation study results is provided in \Cref{appn:additional_ablations}.

\section{Conclusion and Discussion}
\label{sec:conclusion}
\vspace{0.15cm}

We introduce \emph{SphereDiff}, a tuning-free framework for 360$\degree$ live wallpaper generation that effectively leverages state-of-the-art image and video diffusion models.
The proposed spherical latent representation inherently supports consistent generation quality across all view directions, including near the poles.
Our extended MultiDiffusion framework for 360$\degree$ panoramas with dynamic latent sampling facilitates a tuning-free approach.
Lastly, distortion-aware weighted averaging significantly enhances the quality of panoramic content.
In summary, we achieve state-of-the-art performance in generating 360$\degree$ live and static wallpapers, as demonstrated through comprehensive experiments.

\paragraph{Limitations.}
Our approach achieves strong results in producing high-quality static and live 360$\degree$ wallpapers for a broad range of outdoor.
However, it cannot yet generate highly complex scenes, such as indoor environments, without additional training.
Recent panoramic image generation approaches~\cite{kalischek2025cubediff,ccapuk2025tandit} handle more complex scenes by relying on large-scale panoramic image datasets, but they cannot be readily adapted to panoramic videos due to the limited availability of panoramic video datasets.
In contrast, our approach achieves state-of-the-art performance in generating 360$\degree$ live wallpapers without additional tuning, highlighting its potential as a strong foundation for future work on more complex 360$\degree$ panoramic video generation.

\section*{Acknowledgments}

This work was supported by Institute for Information \& communications Technology Planning \& Evaluation(IITP) grant funded by the Korea government(MSIT) (RS-2019-II190075, Artificial Intelligence Graduate School Program(KAIST)).
This work was supported by the National Research Foundation of Korea(NRF) grant funded by the Korea government(MSIT) (No. RS-2025-00555621)
This work was supported by Electronics and Telecommunications Research Institute(ETRI) grant funded by the Korean government [25ZB1200, Fundamental Technology Research for Human-Centric Autonomous Intelligent Systems]

\balance

\bibliography{aaai2026}

@String(CVPR= {IEEE Conf. Comput. Vis. Pattern Recog.})

@String(TOG= {ACM Trans. Graph.})

@String(AAAI = {AAAI})

@String(CVPR  = {CVPR})

@String(TOG   = {ACM TOG})

@article{gpt4o,
  title={Gpt-4o system card},
  author={Hurst, Aaron and Lerer, Adam and Goucher, Adam P and Perelman, Adam and Ramesh, Aditya and Clark, Aidan and Ostrow, AJ and Welihinda, Akila and Hayes, Alan and Radford, Alec and others},
  journal={arXiv preprint arXiv:2410.21276},
  year={2024}
}

@article{guo2023animatediff,
  title={Animatediff: Animate your personalized text-to-image diffusion models without specific tuning},
  author={Guo, Yuwei and Yang, Ceyuan and Rao, Anyi and Liang, Zhengyang and Wang, Yaohui and Qiao, Yu and Agrawala, Maneesh and Lin, Dahua and Dai, Bo},
  journal={arXiv preprint arXiv:2307.04725},
  year={2023}
}

@inproceedings{wang2024360dvd,
  title={360dvd: Controllable panorama video generation with 360-degree video diffusion model},
  author={Wang, Qian and Li, Weiqi and Mou, Chong and Cheng, Xinhua and Zhang, Jian},
  booktitle={Proceedings of the IEEE/CVF Conference on Computer Vision and Pattern Recognition},
  pages={6913--6923},
  year={2024}
}

@article{liu2024dynamicscaler,
  title={DynamicScaler: Seamless and Scalable Video Generation for Panoramic Scenes},
  author={Liu, Jinxiu and Lin, Shaoheng and Li, Yinxiao and Yang, Ming-Hsuan},
  journal={arXiv preprint arXiv:2412.11100},
  year={2024}
}

@article{li20244k4dgen,
  title={4k4dgen: Panoramic 4d generation at 4k resolution},
  author={Li, Renjie and Pan, Panwang and Yang, Bangbang and Xu, Dejia and Zhou, Shijie and Zhang, Xuanyang and Li, Zeming and Kadambi, Achuta and Wang, Zhangyang and Tu, Zhengzhong and others},
  journal={arXiv preprint arXiv:2406.13527},
  year={2024}
}

@inproceedings{kalischek2025cubediff,
  title={Cubediff: Repurposing diffusion-based image models for panorama generation},
  author={Kalischek, Nikolai and Oechsle, Michael and Manhardt, Fabian and Henzler, Philipp and Schindler, Konrad and Tombari, Federico},
  booktitle={The Thirteenth International Conference on Learning Representations},
  year={2025}
}

@inproceedings{zhang2024panfusion,
  title={Taming Stable Diffusion for Text to 360 Panorama Image Generation},
  author={Zhang, Cheng and Wu, Qianyi and Gambardella, Camilo Cruz and Huang, Xiaoshui and Phung, Dinh and Ouyang, Wanli and Cai, Jianfei},
  booktitle={Proceedings of the IEEE/CVF Conference on Computer Vision and Pattern Recognition},
  pages={6347--6357},
  year={2024}
}

@article{chen2022text2light,
  title={Text2light: Zero-shot text-driven hdr panorama generation},
  author={Chen, Zhaoxi and Wang, Guangcong and Liu, Ziwei},
  journal={ACM Transactions on Graphics (TOG)},
  volume={41},
  number={6},
  pages={1--16},
  year={2022},
  publisher={ACM New York, NY, USA}
}

@inproceedings{chang2024warped,
  title={How I Warped Your Noise: a Temporally-Correlated Noise Prior for Diffusion Models},
  author={Chang, Pascal and Tang, Jingwei and Gross, Markus and Azevedo, Vinicius C},
  booktitle={The Twelfth International Conference on Learning Representations},
  year={2024}
}

@article{hacohen2024ltx,
  title={Ltx-video: Realtime video latent diffusion},
  author={HaCohen, Yoav and Chiprut, Nisan and Brazowski, Benny and Shalem, Daniel and Moshe, Dudu and Richardson, Eitan and Levin, Eran and Shiran, Guy and Zabari, Nir and Gordon, Ori and others},
  journal={arXiv preprint arXiv:2501.00103},
  year={2024}
}

@article{yang2024cogvideox,
  title={Cogvideox: Text-to-video diffusion models with an expert transformer},
  author={Yang, Zhuoyi and Teng, Jiayan and Zheng, Wendi and Ding, Ming and Huang, Shiyu and Xu, Jiazheng and Yang, Yuanming and Hong, Wenyi and Zhang, Xiaohan and Feng, Guanyu and others},
  journal={arXiv preprint arXiv:2408.06072},
  year={2024}
}

@article{kong2024hunyuanvideo,
  title={Hunyuanvideo: A systematic framework for large video generative models},
  author={Kong, Weijie and Tian, Qi and Zhang, Zijian and Min, Rox and Dai, Zuozhuo and Zhou, Jin and Xiong, Jiangfeng and Li, Xin and Wu, Bo and Zhang, Jianwei and others},
  journal={arXiv preprint arXiv:2412.03603},
  year={2024}
}

@misc{genmo2024mochi,
      title={Mochi 1},
      author={Genmo Team},
      year={2024},
      publisher = {GitHub},
      journal = {GitHub repository},
      howpublished={\url{https://github.com/genmoai/models}}
}

@article{zhang2025flashvideo,
  title={FlashVideo: Flowing Fidelity to Detail for Efficient High-Resolution Video Generation},
  author={Zhang, Shilong and Li, Wenbo and Chen, Shoufa and Ge, Chongjian and Sun, Peize and Zhang, Yida and Jiang, Yi and Yuan, Zehuan and Peng, Binyue and Luo, Ping},
  journal={arXiv preprint arXiv:2502.05179},
  year={2025}
}

@article{liu2025lumina,
  title={Lumina-Video: Efficient and Flexible Video Generation with Multi-scale Next-DiT},
  author={Liu, Dongyang and Li, Shicheng and Liu, Yutong and Li, Zhen and Wang, Kai and Li, Xinyue and Qin, Qi and Liu, Yufei and Xin, Yi and Li, Zhongyu and others},
  journal={arXiv preprint arXiv:2502.06782},
  year={2025}
}

@inproceedings{huang2024vbench,
  title={Vbench: Comprehensive benchmark suite for video generative models},
  author={Huang, Ziqi and He, Yinan and Yu, Jiashuo and Zhang, Fan and Si, Chenyang and Jiang, Yuming and Zhang, Yuanhan and Wu, Tianxing and Jin, Qingyang and Chanpaisit, Nattapol and others},
  booktitle={Proceedings of the IEEE/CVF Conference on Computer Vision and Pattern Recognition},
  pages={21807--21818},
  year={2024}
}

@article{Peebles2022DiT,
  title={Scalable Diffusion Models with Transformers},
  author={William Peebles and Saining Xie},
  year={2022},
  journal={arXiv preprint arXiv:2212.09748},
}

@inproceedings{ldm,
  title={High-resolution image synthesis with latent diffusion models},
  author={Rombach, Robin and Blattmann, Andreas and Lorenz, Dominik and Esser, Patrick and Ommer, Bj{\"o}rn},
  booktitle={Proceedings of the IEEE/CVF conference on computer vision and pattern recognition},
  pages={10684--10695},
  year={2022}
}

@article{xie2024sana,
  title={Sana: Efficient high-resolution image synthesis with linear diffusion transformers},
  author={Xie, Enze and Chen, Junsong and Chen, Junyu and Cai, Han and Tang, Haotian and Lin, Yujun and Zhang, Zhekai and Li, Muyang and Zhu, Ligeng and Lu, Yao and others},
  journal={arXiv preprint arXiv:2410.10629},
  year={2024}
}

@misc{flux2024,
    author={Black Forest Labs},
    title={FLUX},
    year={2024},
    howpublished={\url{https://github.com/black-forest-labs/flux}},
}

@inproceedings{bar2023multidiffusion,
  title={MultiDiffusion: fusing diffusion paths for controlled image generation},
  author={Bar-Tal, Omer and Yariv, Lior and Lipman, Yaron and Dekel, Tali},
  booktitle={Proceedings of the 40th International Conference on Machine Learning},
  pages={1737--1752},
  year={2023}
}

@article{chen2023pixart,
  title={Pixart-$\alpha$: Fast training of diffusion transformer for photorealistic text-to-image synthesis},
  author={Chen, Junsong and Yu, Jincheng and Ge, Chongjian and Yao, Lewei and Xie, Enze and Wu, Yue and Wang, Zhongdao and Kwok, James and Luo, Ping and Lu, Huchuan and others},
  journal={arXiv preprint arXiv:2310.00426},
  year={2023}
}

@article{fibonacci,
  title={A Comparison of Popular Point Configurations on $\mathbb{S}^2$},
  author={Hardin, Doug P and Michaels, TJ and Saff, Edward B},
  journal={arXiv preprint arXiv:1607.04590},
  year={2016}
}

@misc{360_lora,
    author={LatentLabs360},
    title={LatentLabs360},
    year={2023},
    publisher = {CivitAI},
    howpublished={\url{https://civitai.com/models/10753/latentlabs360}},
}

@article{zheng2023judging,
  title={Judging llm-as-a-judge with mt-bench and chatbot arena},
  author={Zheng, Lianmin and Chiang, Wei-Lin and Sheng, Ying and Zhuang, Siyuan and Wu, Zhanghao and Zhuang, Yonghao and Lin, Zi and Li, Zhuohan and Li, Dacheng and Xing, Eric and others},
  journal={Advances in neural information processing systems},
  volume={36},
  pages={46595--46623},
  year={2023}
}

@inproceedings{heusel2017gans,
  title     = {GANs Trained by a Two Time-Scale Update Rule Converge to a Local Nash Equilibrium},
  author    = {Heusel, Martin and Ramsauer, Hubert and Unterthiner, Thomas and Nessler, Bernhard and Hochreiter, Sepp},
  booktitle = {Advances in Neural Information Processing Systems},
  pages     = {6626--6637},
  year      = {2017}
}

@article{ccapuk2025tandit,
  title={TanDiT: Tangent-Plane Diffusion Transformer for High-Quality 360 $\{$$\backslash$deg$\}$ Panorama Generation},
  author={{\c{C}}apuk, Hakan and Bond, Andrew and K{\i}z{\i}l, Muhammed Burak and G{\"o}{\c{c}}en, Emir and Erdem, Erkut and Erdem, Aykut},
  journal={arXiv preprint arXiv:2506.21681},
  year={2025}
}

@article{lu2024genex,
  title={Genex: Generating an explorable world},
  author={Lu, Taiming and Shu, Tianmin and Xiao, Junfei and Ye, Luoxin and Wang, Jiahao and Peng, Cheng and Wei, Chen and Khashabi, Daniel and Chellappa, Rama and Yuille, Alan and others},
  journal={arXiv preprint arXiv:2412.09624},
  year={2024}
}

@article{wei2022chain,
  title={Chain-of-thought prompting elicits reasoning in large language models},
  author={Wei, Jason and Wang, Xuezhi and Schuurmans, Dale and Bosma, Maarten and Xia, Fei and Chi, Ed and Le, Quoc V and Zhou, Denny and others},
  journal={Advances in neural information processing systems},
  volume={35},
  pages={24824--24837},
  year={2022}
}

@inproceedings{you2024depicting,
  title={Depicting beyond scores: Advancing image quality assessment through multi-modal language models},
  author={You, Zhiyuan and Li, Zheyuan and Gu, Jinjin and Yin, Zhenfei and Xue, Tianfan and Dong, Chao},
  booktitle={European Conference on Computer Vision},
  pages={259--276},
  year={2024},
  organization={Springer}
}

@inproceedings{wang2025your,
  title={Is your world simulator a good story presenter? a consecutive events-based benchmark for future long video generation},
  author={Wang, Yiping and He, Xuehai and Wang, Kuan and Ma, Luyao and Yang, Jianwei and Wang, Shuohang and Du, Simon Shaolei and Shen, Yelong},
  booktitle={Proceedings of the Computer Vision and Pattern Recognition Conference},
  pages={13629--13638},
  year={2025}
}

@article{you2024descriptive,
  title={Descriptive image quality assessment in the wild},
  author={You, Zhiyuan and Gu, Jinjin and Li, Zheyuan and Cai, Xin and Zhu, Kaiwen and Dong, Chao and Xue, Tianfan},
  journal={arXiv preprint arXiv:2405.18842},
  year={2024}
}

@article{zhou2024vision,
  title={Vision language modeling of content, distortion and appearance for image quality assessment},
  author={Zhou, Fei and Gu, Tianhao and Huang, Zhicong and Qiu, Guoping},
  journal={IEEE Journal of Selected Topics in Signal Processing},
  year={2024},
  publisher={IEEE}
}

@article{daras2024warped,
  title={Warped diffusion: Solving video inverse problems with image diffusion models},
  author={Daras, Giannis and Nie, Weili and Kreis, Karsten and Dimakis, Alex and Mardani, Morteza and Kovachki, Nikola and Vahdat, Arash},
  journal={Advances in Neural Information Processing Systems},
  volume={37},
  pages={101116--101143},
  year={2024}
}

@article{Matterport3D,
  title={Matterport3D: Learning from RGB-D Data in Indoor Environments},
  author={Chang, Angel and Dai, Angela and Funkhouser, Thomas and Halber, Maciej and Niessner, Matthias and Savva, Manolis and Song, Shuran and Zeng, Andy and Zhang, Yinda},
  journal={International Conference on 3D Vision (3DV)},
  year={2017}
}

@inproceedings{zheng2020structured3d,
  title={Structured3d: A large photo-realistic dataset for structured 3d modeling},
  author={Zheng, Jia and Zhang, Junfei and Li, Jing and Tang, Rui and Gao, Shenghua and Zhou, Zihan},
  booktitle={European Conference on Computer Vision},
  pages={519--535},
  year={2020},
  organization={Springer}
}

@article{feng2023diffusion360,
  title={Diffusion360: Seamless 360 degree panoramic image generation based on diffusion models},
  author={Feng, Mengyang and Liu, Jinlin and Cui, Miaomiao and Xie, Xuansong},
  journal={arXiv preprint arXiv:2311.13141},
  year={2023}
}

@article{mvdiffusion,
  title={Emergent correspondence from image diffusion},
  author={Tang, Luming and Jia, Menglin and Wang, Qianqian and Phoo, Cheng Perng and Hariharan, Bharath},
  journal={Advances in Neural Information Processing Systems},
  volume={36},
  pages={1363--1389},
  year={2023}
}

@article{barron2022mipnerf360,
    title={Mip-NeRF 360: Unbounded Anti-Aliased Neural Radiance Fields},
    author={Jonathan T. Barron and Ben Mildenhall and 
            Dor Verbin and Pratul P. Srinivasan and Peter Hedman},
    journal={CVPR},
    year={2022}
}

@article{liao2023cylin,
  title={Cylin-painting: Seamless 360 panoramic image outpainting and beyond},
  author={Liao, Kang and Xu, Xiangyu and Lin, Chunyu and Ren, Wenqi and Wei, Yunchao and Zhao, Yao},
  journal={IEEE Transactions on Image Processing},
  volume={33},
  pages={382--394},
  year={2023},
  publisher={IEEE}
}

@inproceedings{wu2024panodiffusion,
  title={PanoDiffusion: 360-degree Panorama Outpainting via Diffusion},
  author={Wu, Tianhao and Zheng, Chuanxia and Cham, Tat-Jen},
  year={2024},
  booktitle={The Twelfth International Conference on Learning Representations}
}

@inproceedings{wu2024spherediffusion,
  title={Spherediffusion: Spherical geometry-aware distortion resilient diffusion model},
  author={Wu, Tao and Li, Xuewei and Qi, Zhongang and Hu, Di and Wang, Xintao and Shan, Ying and Li, Xi},
  booktitle={Proceedings of the AAAI Conference on Artificial Intelligence},
  pages={6126--6134},
  year={2024}
}



\setcounter{section}{0}
\maketitlesupplementary

\nobalance

\renewcommand{\thesection}{\Alph{section}}

In the appendix, we include method descriptions, comparisons, ablations, and evaluation details.
First, we present method details with mathematical expressions and visual illustrations for completeness (\Cref{appn:spherical_to_perspective,appn:multi_prompt}).
Next, we present in-depth ablation studies and detailed comparisons with baselines, including additional qualitative and quantitative results (\Cref{appn:additional_ablations,appn:comparison}).
Finally, we provide implementation and evaluation details, along with the text prompts used to generate the results (\Cref{appn:implementation_details,,appn:evaluation_details,,appn:text_prompts}).
The table of contents for our appendix is provided below.

\section*{Table of Content}

\vspace{0.1cm}
\begin{enumerate}[label=\textbf{\Alph*.}]
    \item \textbf{\nameref{appn:spherical_to_perspective}}
    \begin{enumerate}[label=\Alph{enumi}.\arabic*.]
        \item \nameref{appn:spherical_to_perspective_mapping}
        \item \nameref{appn:spherical_to_perspective_distortion}
        \item \nameref{appn:spherical_to_perspective_coord}
    \end{enumerate}
    \item \textbf{\nameref{appn:multi_prompt}}
    \begin{enumerate}[label=\Alph{enumi}.\arabic*.]
        \item \nameref{appn:multi_prompt_visualization}
        \item \nameref{appn:multi_prompt_foreground}
    \end{enumerate}
    \item \textbf{\nameref{appn:additional_ablations}}
    \begin{enumerate}[label=\Alph{enumi}.\arabic*.]
        \item \nameref{appn:ablation_quanti}
        \item \nameref{appn:ablation_alternatives}
    \end{enumerate}
    \item \textbf{\nameref{appn:comparison}}
    \begin{enumerate}[label=\Alph{enumi}.\arabic*.]
        \item \nameref{appn:comparison_quali}
        \item \nameref{appn:comparison_baselines}
    \end{enumerate}
    \item \textbf{\nameref{appn:implementation_details}}
    \item \textbf{\nameref{appn:evaluation_details}}
    \begin{enumerate}[label=\Alph{enumi}.\arabic*.]
        \item \nameref{appn:evaluation_user_study}
        \item \nameref{appn:evaluation_vlm}
    \end{enumerate}
    \item \textbf{\nameref{appn:text_prompts}}
\end{enumerate}
\vspace{0.1cm}

\section{Spherical-to-Perspective Projection}
\label{appn:spherical_to_perspective}

In this section, we describe the formulation of the spherical-to-projection mapping functions (\Cref{appn:spherical_to_perspective_mapping}) and provide further explanation of the spherical-to-perspective distortion (\Cref{appn:spherical_to_perspective_distortion}) for completeness.

\subsection{Spherical-to-Perspective Mapping Function}
\label{appn:spherical_to_perspective_mapping}

For perspective latent representation, we define a virtual camera centered at the origin.
The points in world coordinates, denoted as $ d = (x, y, z)^\top \in \mathbb{S}^2$, are projected onto image space using the projection matrix $P = K[R|t]$. Here, $K$ is the intrinsic camera matrix derived from a predefined focal length $f$, $R$  represents the viewing direction, and $t$ is set to zero.  
The spherical-to-perspective projection function $\mathcal{T}_{\mathbb{S}^2 \rightarrow \mathbb{P}^2}$ can be formulated as:
\begin{equation}
    \tilde{u} = K [R | t] \tilde{d},
\end{equation}
where $ \tilde{d} = (x, y, z, 1)^\top $ is the homogeneous coordinate representation of the 3D point $d$, and $ \tilde{u} = (u', v', w')^\top $ represents the projected homogeneous coordinates in image space. The final 2D perspective coordinates $ u = (u, v)^\top \in \mathbb{P}^2$ are obtained via perspective division:
\begin{equation}
    u = \left( \frac{u'}{w'}, \frac{v'}{w'} \right).
\end{equation}
To ensure proper visibility, points located behind the view direction are masked out using their inner product values, retaining only the points in the frontal view.

\begin{figure}[t]
\centering
\includegraphics[width=1.0\columnwidth]{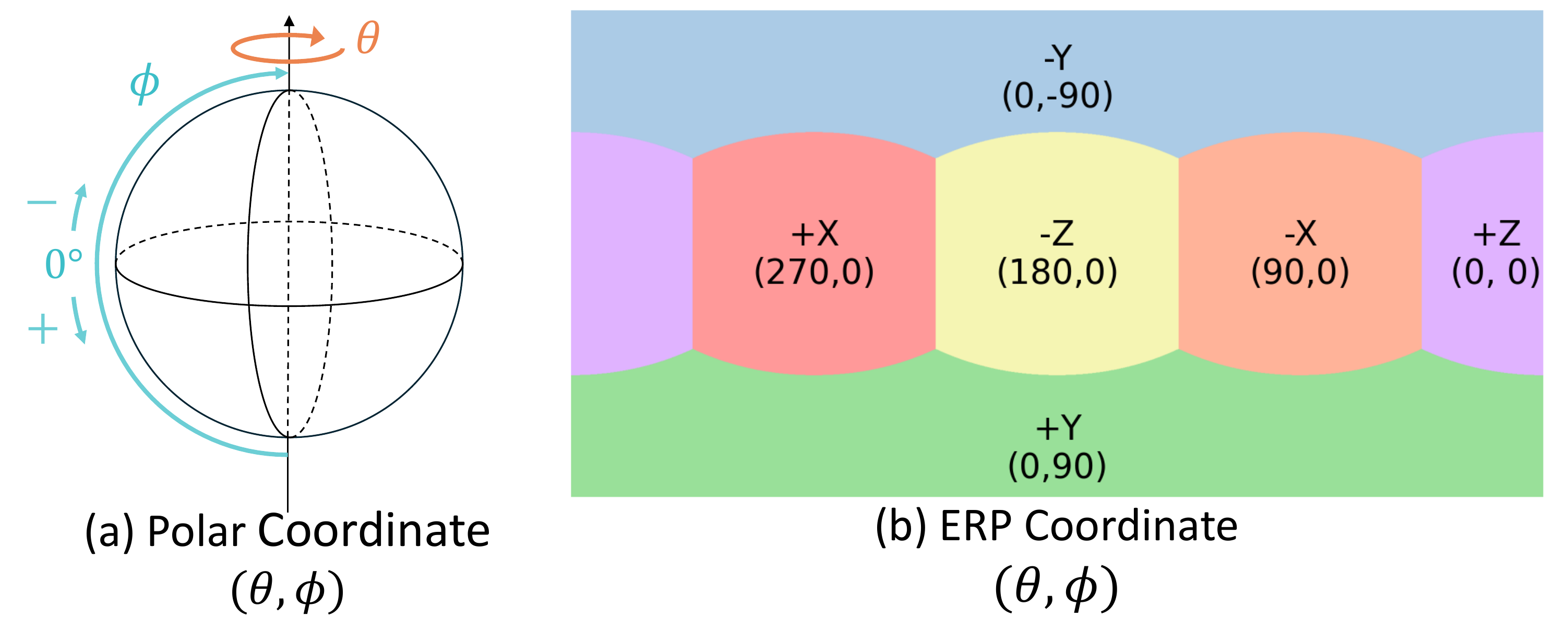}
\caption{
\textbf{Our Coordinate System.}
(a) The spherical coordinate system with azimuth $\theta$ and elevation $\phi$. (b) The corresponding layout in the equirectangular projection (ERP) format, showing the mapping of Cartesian axes to spherical coordinates $(\theta, \phi)$.
}
\label{fig:A_polar_coordinate}
\end{figure}

\begin{figure*}[t]
\centering
\includegraphics[width=\linewidth]{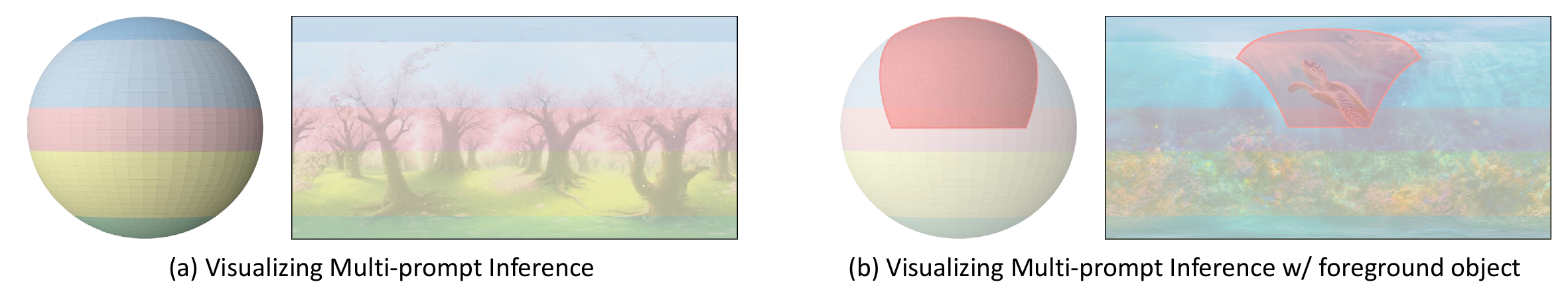}
\caption{
\textbf{Visual Illustration of Multi-Prompt Inference.}  
We illustrate the multi-prompt inference process using visual aids.
(a) Three different prompts are assigned based on the elevation angle ($\phi$), as indicated by the color-coded regions in both the spherical projection and ERP, for generating 360$\degree$ wallpapers.
(b) A foreground object is generated using an additional foreground prompt ($\vy_\text{foreground}$) conditioned on the corresponding view direction ($\theta = 180\degree, \phi = -40\degree$).
}
\label{fig:A_multi_prompt_inference}
\end{figure*}

\subsection{Spherical-to-Perspective Distortion}
\label{appn:spherical_to_perspective_distortion}

The core methodology of our model involves approximating the spherical surface with a local tangent plane for computational purposes. However, this approach inherently introduces a form of geometric distortion. The tangent plane approximation is highly accurate at its central point, which corresponds to the view direction, but the deviation between the true spherical surface and the approximated plane systematically increases with the radial distance from this center. Consequently, a greater degree of error accumulates toward the periphery of the image.

\paragraph{Geometric Formulation.}

To formalize this, let us consider a sphere of radius $R$ centered at the origin $(0, 0, 0)$. We define a tangent plane to be in contact with the sphere at its north pole, $P_N = (0, 0, R)$. The equation of this plane is therefore $z=R$.
A point $P_s$ on the surface of the sphere can be described by its angle $\theta$ from the positive z-axis. For simplicity, we can analyze the geometry in a 2D cross-section. In this view, the coordinates of the point $P_s$ are $(R\sin\theta, R\cos\theta)$.
The projection ray originates from the center of the sphere $(0,0,0)$ and passes through $P_s$ to intersect the tangent plane ($z=R$) at a projected point, $P_p$. Let the distance of this projected point from the center of the plane be $d_p$. The coordinates of $P_p$ are thus $(d_p, R)$ in the 2D cross-section.
By the property of similar triangles formed by the origin, the z-axis, and the projection ray, the ratio of the horizontal to the vertical component is constant:

\[
\frac{\text{horizontal}}{\text{vertical}} = \frac{R\sin\theta}{R\cos\theta} = \frac{d_p}{R}
\]

This simplifies to $\tan\theta = d_p/R$. Solving for the projected distance $d_p$, we get:

\[
d_p = R \tan\theta
\]

\paragraph{Analysis of Distortion.}

The distortion arises from the non-linear relationship between the true distance on the sphere's surface and the projected distance on the plane.

\begin{enumerate}
    \item \textbf{True Surface Distance:} The actual distance from the pole $P_N$ to the point $P_s$ along the arc of the sphere is the arc length, $d_s$, given by:
    \[
    d_s = R\theta \quad (\text{where } \theta \text{ is in radians})
    \]

    \item \textbf{Projected Distance:} As derived above, the distance from the center of the plane to the projected point $P_p$ is:
    \[
    d_p = R \tan\theta
    \]
\end{enumerate}

Comparing these two distances reveals the nature of the distortion. For small angles near the center of projection ($\theta \approx 0$), the approximation $\tan\theta \approx \theta$ holds, which means $d_p \approx d_s$. This indicates that there is minimal distortion near the point of tangency (i.e., the view direction).
However, as $\theta$ increases, the value of $\tan\theta$ grows much more rapidly than $\theta$. This non-linear relationship, $d_p/d_s = \tan\theta / \theta > 1$, causes a radial stretching effect. Consequently, features that are uniformly spaced on the spherical surface become increasingly spread out as they are projected farther from the center of the plane. This effect is precisely the spherical-to-perspective distortion observed the left image of \Cref{fig:3_discretization}, where the periphery of the image appears magnified relative to the center. 

To mitigate this distortion, we employ a weighted average with a standard exponential function. While this weighting scheme is not a perfect theoretical inverse of the distortion, our empirical results indicate that more complex models do not yield a significant improvement in quality. Therefore, we just adopted the simplest exponential form, $ W^{jk}_i = \exp \left( - \norm{\vu_{jk}} / \tau \right)$.

\subsection{Coordinate Systems.}
\label{appn:spherical_to_perspective_coord}

We employ a left-handed Cartesian coordinate system where the $+X$, $+Y$, and $+Z$ axes point right, down, and forward (towards the camera), respectively. For panoramic representation, we use a spherical coordinate system defined by an azimuth angle $\theta \in [0\degree, 360\degree)$ and an elevation angle $\phi \in [-90\degree, 90\degree]$.
As illustrated in~\Cref{fig:A_polar_coordinate}, the azimuth $\theta$ represents the horizontal rotation, starting from the $+Z$ axis and increasing clockwise. The elevation $\phi$ represents the vertical angle from the horizontal plane; a value of $\phi=0\degree$ corresponds to the horizon, while negative and positive values correspond to the upper (upward-looking) and lower (downward-looking) hemispheres, respectively.

\begin{table*}[t]
\centering
\resizebox{0.85\linewidth}{!}{
\begin{tabular}{l|c|cccccc}

\toprule
& \multirow{2}{*}[-0.3cm]{\hspace{-0.15cm}\begin{tabular}{c}Runtime \\ (s)\end{tabular}\hspace{-0.15cm}} & \multicolumn{2}{c}{Panoramic Criteria} & \multicolumn{2}{c}{Image Criteria} & \multicolumn{2}{c}{Text Adherence} \\
\cmidrule(lr){3-4} \cmidrule(lr){5-6} \cmidrule(lr){7-8}
Method &  & Distortion $\uparrow$ & End Continuity $\uparrow$ & \begin{tabular}[c]{@{}c@{}}Image\\ Quality\end{tabular} $\uparrow$ & \begin{tabular}[c]{@{}c@{}}Aesthetic\\ Appearance\end{tabular} $\uparrow$ & Scene $\uparrow$ & CLIP-Score $\uparrow$ \\
\drule
Nearest sampling & 161 & 2.039 & 3.400 & 3.014 & 4.057 & 0.3250 & 27.19\\
+ Weighted Avg.& 162 & 2.829 & 4.625 & 4.139 & \textbf{4.782} & 0.5125 & 28.26 \\ \midrule
Dynamic sampling & 184 & 2.421 & 4.086 & 3.454 & 4.111 & 0.1375 & 25.92 \\
+ Weighted Avg. & 185 & \textbf{3.238} & \textbf{4.892} & \textbf{4.496} & 4.685 & \textbf{0.5875} & \textbf{28.65} \\
\bottomrule

\end{tabular}
}
\caption{
Automated quantitative ablation study in generating 360$\degree$ static wallpaper generation (SANA, A100-40GB) including runtime.
Dynamic latent sampling improves distortion and end-continuity, while the distortion-aware weighted averaging significantly improves the image quality and text adherence.
}
\label{tab:A_ablation}
\end{table*}

\section{Details of Multi-Prompt Inference}
\label{appn:multi_prompt}

We further describe the multi-prompt inference method in \Cref{sec:3_method_multi_prompt}, accompanied by a visual illustration (\Cref{appn:multi_prompt_visualization}), and provide additional details on foreground object generation (\Cref{appn:multi_prompt_foreground}).

\subsection{Visual Illustration of Multi-prompt Inference}
\label{appn:multi_prompt_visualization}

Unlike standard images and videos, 360$\degree$ panoramic content exhibits substantial variation depending on spatial location.
For instance, as shown in \Cref{fig:A_multi_prompt_inference} (a), a 360$\degree$ panorama exhibits significant variation depending on the vertical position in the ERP representation—lower regions often depict the ground, middle regions show elements like blossoms, and upper regions correspond to the sky.
To address this, we provide three distinct prompts corresponding to different elevation levels of the spherical latents.
A visual illustration of the condition mapping function is shown in both spherical and ERP formats in \Cref{fig:A_multi_prompt_inference}, where each color represents a distinct prompt.  
Examples of the prompts are listed in \Cref{appn:text_prompts} and visualized in \Cref{fig:A_used_prompts}.
For a fair comparison, we apply the same multi-prompt inference strategy and prompt settings to the applicable baseline method~\cite{liu2024dynamicscaler}.

\subsection{Foreground Generation}
\label{appn:multi_prompt_foreground}

To synthesize more complex scenes, our method incorporates additional view directions specifically designated for foreground subjects.
While the core generation pipeline remains unchanged, it is augmented with one or two foreground-specific views, as illustrated in \Cref{fig:A_multi_prompt_inference}(b).
For example, a turtle is added as a foreground object in the red-highlighted region.

To ensure that the foreground prompt dominates within the specified region and achieves a clearer separation between the subject and the background, we replace the exponential weighting function with a beta kernel, applied only to the foreground prompt.
The weight is defined as:
\begin{align}
W^{jk}_i &= \mathcal{B}(\|\vu^*_{jk}\|; b) \\
\mathcal{B}(x; b) &= (1 - x)^{4e^b}, \quad \text{where} \,\, x \in [0, 1],\, b \in \mathbb{R}.
\end{align}
Here, $\vu^*_{jk}$ is a normalized distance vector, where $x = \min(\|\vu_{jk}\| / d_{\text{max}}, 1)$ and $d_{\text{max}}$ denotes the maximum distance.
The shape of the kernel is controlled by parameter $b$; we set $b = -3.0$ to emphasize the foreground region.
Compared to the exponential kernel, the beta kernel produces a flatter distribution, allowing the model to generate foreground objects that remain dominant throughout the entire foreground region.

\paragraph{Refinement Stage.}

As suggested in previous studies~\cite{liu2024dynamicscaler,ccapuk2025tandit}, we introduce a refinement stage to enhance the visual fidelity and seamlessness of the final output.
After the initial panorama generation, we perform a noise-to-denoise process: noise corresponding to a specific timestep is added to the panorama, which is then refined by a pretrained diffusion model.
This leverages the model’s generative capabilities to improve fine details and overall visual coherence.
The refinement is selectively applied only to panoramas that include foreground elements (\ie, only for SANA in \Cref{fig:1_teaser}) due to the increased computational cost, but it is particularly effective at reducing seams around the foreground-background boundary.

\begin{figure}[t]
\centering
\includegraphics[width=0.95\linewidth]{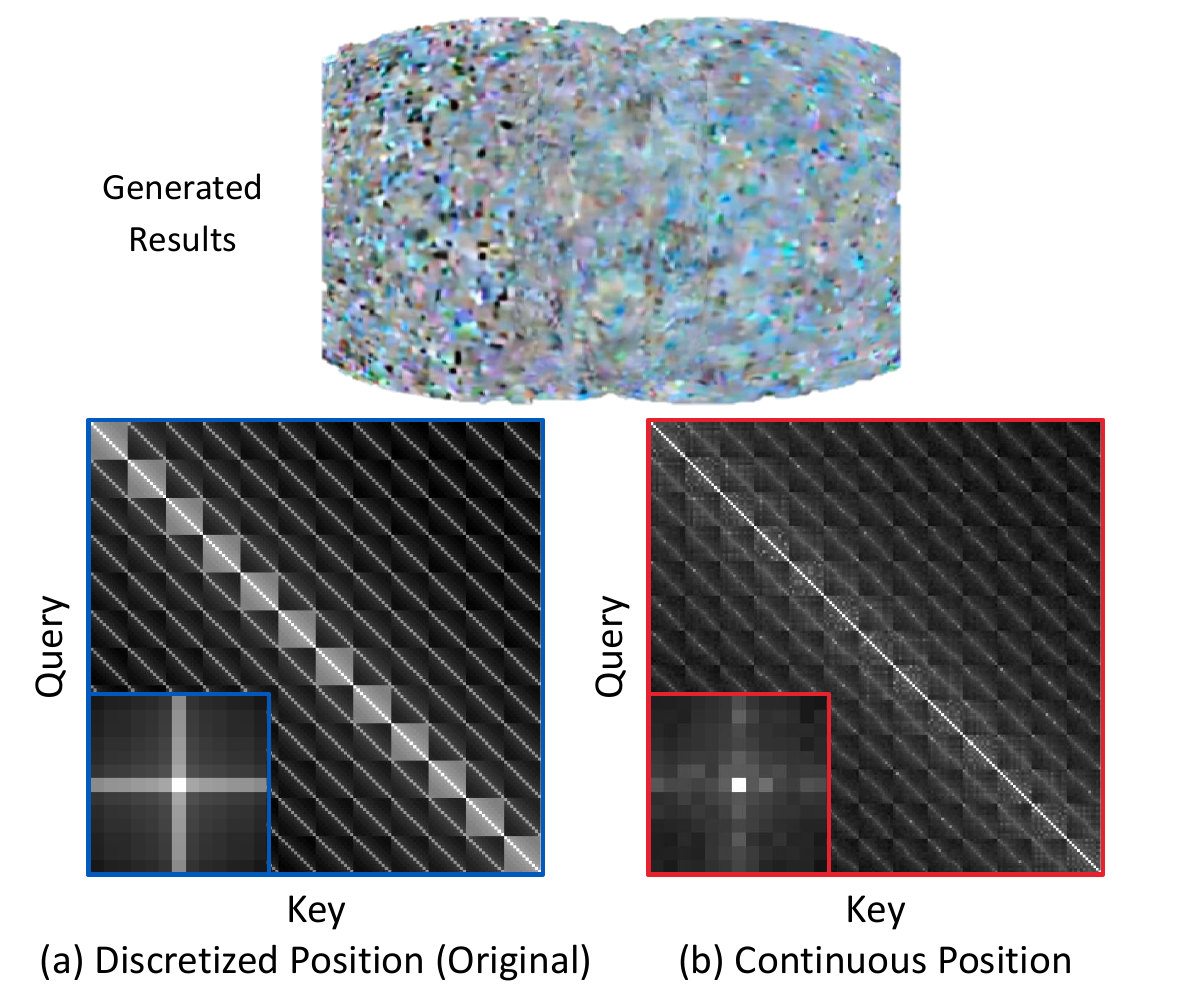}
\caption{
\textbf{Similarities Between Positional Embeddings.}
The small squares represent the similarity when the central pixel is used as the query.
Discretized position originally provides high similarity within the same row or column.
In contrast, even slight variations in continuous position result in a significant drop in similarity.
As a result, it fails to generate reasonable results due to the significant distribution shift.
}
\label{fig:A_pe}
\end{figure}

\begin{figure*}[t]
\centering
\includegraphics[width=\linewidth]{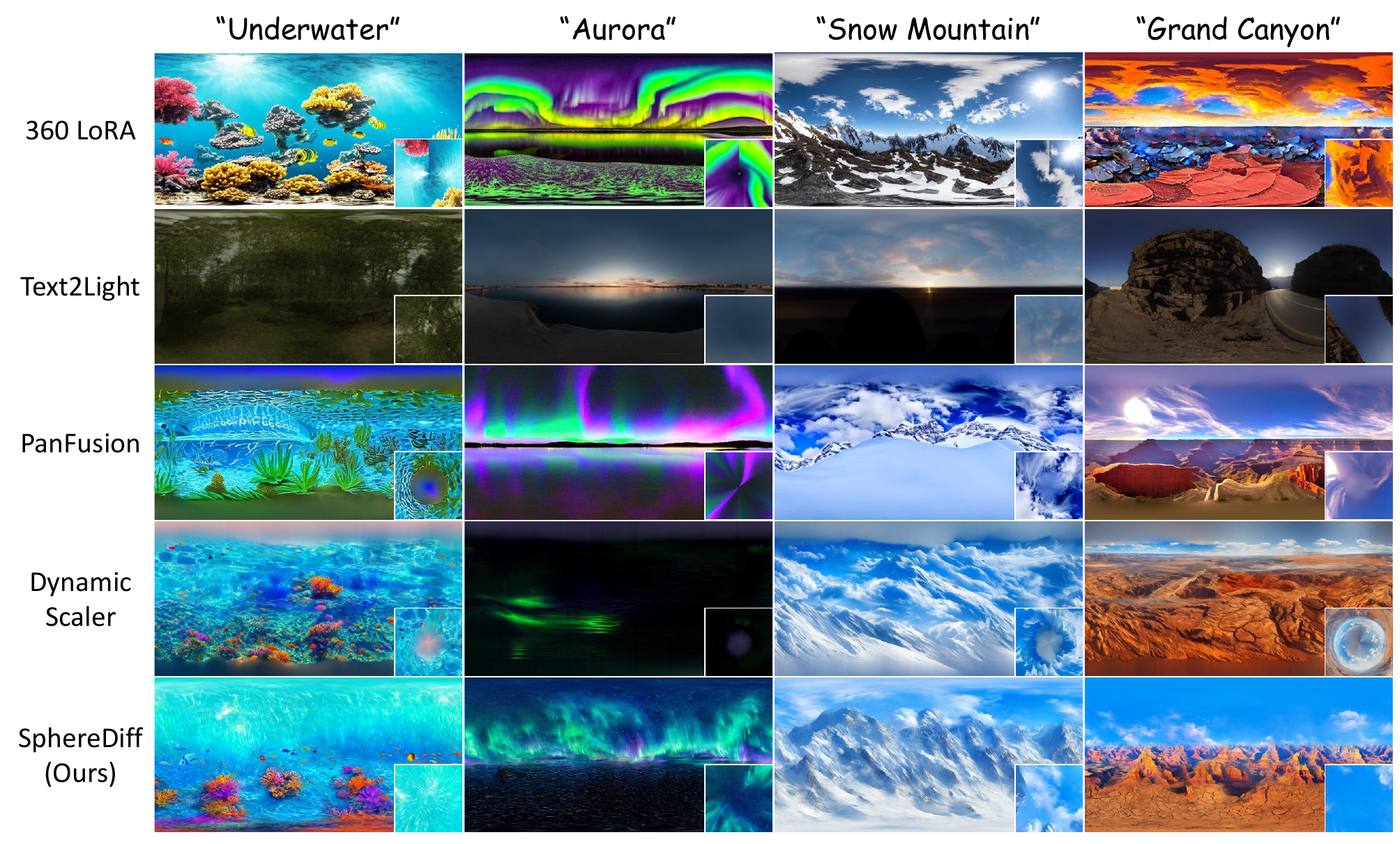}
\caption{
\textbf{Qualitative Comparison with Full ERP (360$\degree$ static wallpapers).}
The first column shows the full ERP views corresponding to \Cref{fig:4_quali_comparison}.
The baselines often fail to generate seamless results near the poles, indicating their inability to produce true 360$\degree$ panoramas.
While Text2Light~\cite{chen2022text2light} generates visually plausible 360$\degree$ wallpapers, it struggles to adhere to the input text prompts due to the limited diversity of its training dataset.
In contrast, our method produces coherent 360$\degree$ panoramas that better reflect the given text conditions.
}
\label{fig:A_full_erp_comparison_t2i}
\end{figure*}

\section{In-depth Ablations}
\label{appn:additional_ablations}

In the following sections, we present the quantitative results of the ablation study in \Cref{sec:exp_ablation}, which were previously shown only visually (\Cref{appn:ablation_quanti}).
We then discuss our exploration of reasonable alternatives for mapping spherical latents and the rationale behind our final choice (\Cref{appn:ablation_alternatives}).

\subsection{Ablation Study with Automated Quantitative Metrics}
\label{appn:ablation_quanti}
As shown in~\Cref{tab:A_ablation}, we conducted a quantitative ablation study on 360$\degree$ static wallpaper generation (SANA) to isolate and measure the impact of our core components: dynamic sampling and weighted average.
Our analysis compares the full model against ablated versions, including a baseline that utilizes nearest point sampling (detailed in \Cref{sec:3_method_sampling}) without a weighted average.
The results indicate that dynamic sampling is crucial for improving panorama quality metrics.
In addition, the weighted average is primarily beneficial for image alignment and text adherence criteria in both nearest and dynamic sampling.
Ultimately, our full model, which combines both techniques, demonstrates the best overall performance on most metrics.
On the aesthetic appearance metric, its performance is comparable, confirming the synergistic benefits of our proposed methods without sacrificing visual quality.

\begin{figure*}[t]
\centering
\includegraphics[width=\linewidth]{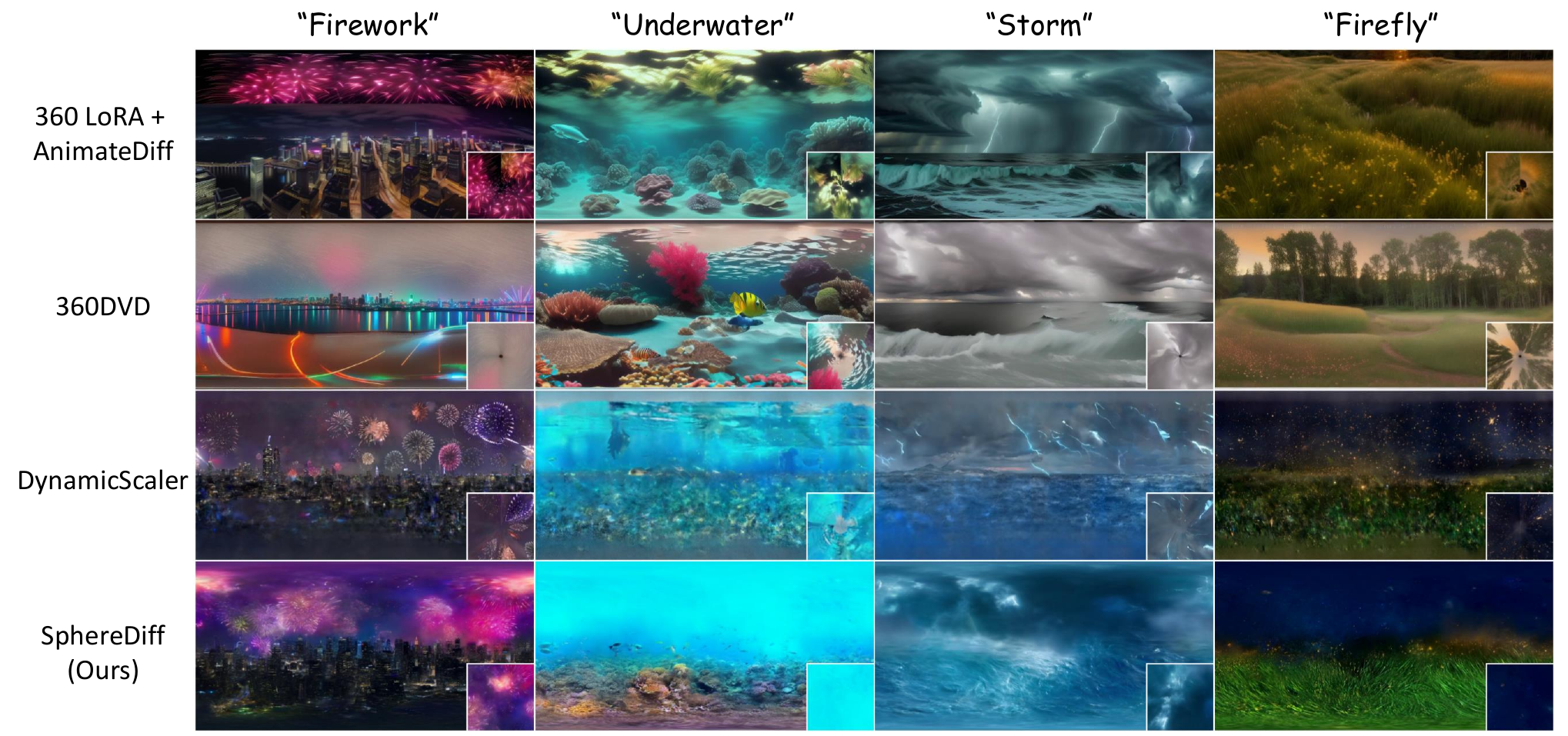}
\caption{
\textbf{Qualitative Comparison with Full ERP (360$\degree$ Live Wallpapers).}  
The last column presents the full ERP views corresponding to \Cref{fig:4_quali_comparison}.  
Compared to our method, baseline approaches often struggle to maintain spatial continuity near the poles, revealing their limitations in achieving seamless 360$\degree$ generation in dynamic settings.  
Please refer to the supplementary material for the full video results.
}
\label{fig:A_full_erp_comparison_t2v}
\end{figure*}

\subsection{Alternatives for Mapping Spherical Latent}
\label{appn:ablation_alternatives}


\paragraph{Directly Use Continuous Position.}

Recent visual generation models, including those based on DiT~\cite{Peebles2022DiT}, support continuous 1D representations through corresponding positional embeddings. A naive approach would be to leverage this property and treat the latent representations as continuous without discretization.
However, this approach leads to unstructured outputs due to a distribution shift in positional embeddings. 
Specifically, in the original positional embedding space, latent similarities are high within the same row or column, ensuring spatial consistency.
In contrast, when using continuous positional embeddings, the similarity between two adjacent points is not necessarily high, even if their spatial coordinates are close, as shown in ~\Cref{fig:A_pe}. 
This discrepancy causes the model to fail in generating structured content. 
Although DiT can process continuous inputs, discretization remains essential for tuning-free panoramic visual generation to maintain structured and consistent latent relationships.

\paragraph{Latent Interpolation Methods.}

To map the spherical latents onto perspective, interpolation is typically used for RGB images.
Conventional interpolation methods, such as bilinear interpolation, when applied in latent space do not provide satisfactory results due to the lack of interpolation-equivariant properties in VAEs.
Thus, stochastic warping methods~\cite{chang2024warped, daras2024warped} have been proposed for warping the latents.
In our experience, we observed that they also provide suboptimal results in generating 360$\degree$ live wallpapers, and thus we adopt sampling methods rather than interpolation methods.
Nevertheless, further studies on improving spherical latent sampling could be a promising direction for 360$\degree$ wallpaper generation.

\section{Detailed Comparison with Baselines}
\label{appn:comparison}

We provide qualitative comparisons with baselines in ERP format (\Cref{appn:comparison_quali}) and present a detailed discussion of related work on 360$\degree$ panorama generation (\Cref{appn:comparison_baselines}).

\subsection{Qualitative Comparison in ERP format}
\label{appn:comparison_quali}

\paragraph{Qualitative ERP Comparison of 360$\degree$ Static Wallpaper Generation Methods.}

We provide full ERP comparisons for 360$\degree$ static and live wallpaper generation methods in \Cref{fig:A_full_erp_comparison_t2i} and \Cref{fig:A_full_erp_comparison_t2v}, respectively.
In the static wallpaper comparison, although other methods appear to generate reasonable panoramas, they often fail to produce seamless 360$\degree$ panoramas under the ERP constraints or to faithfully reflect the input text prompts.
Specifically, as shown in the bottom-right corner, 360 LoRA~\cite{360_lora}, PanFusion~\cite{zhang2024panfusion}, and DynamicScaler~\cite{liu2024dynamicscaler} often fail to generate seamless upper-view perspectives and exhibit distortion, blurriness, and speckling.
While Text2Light~\cite{chen2022text2light} can generate seamless results even for the upper-view perspective, it struggles to produce content that aligns with the given prompts.

\begin{figure*}[!t]
\centering
\includegraphics[width=\linewidth]{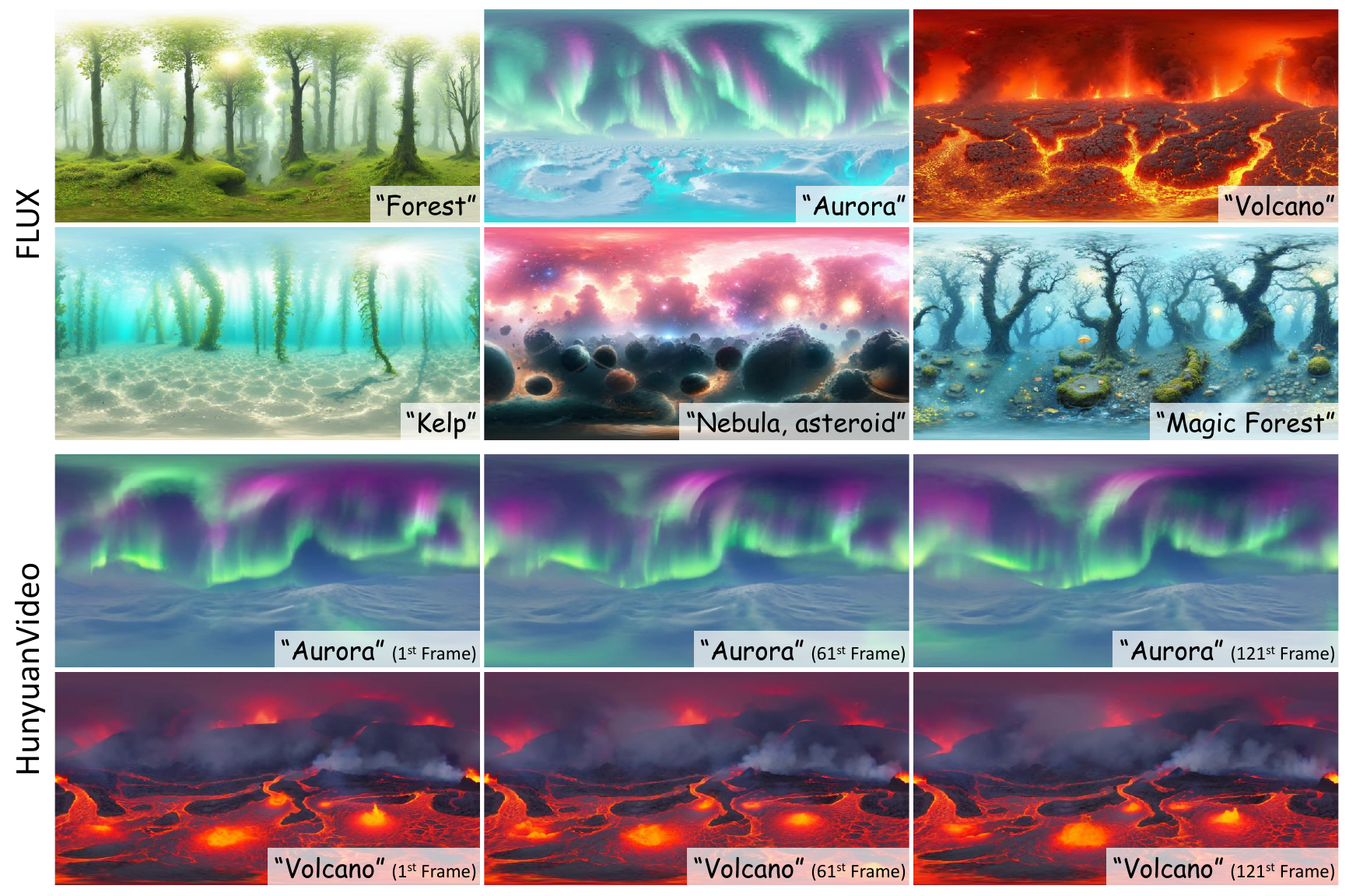}
\caption{
\textbf{Qualitative Results with Different Diffusion Backbones.}  
We extend our method to advanced large-scale diffusion models~\cite{flux2024, kong2024hunyuanvideo}, beyond the efficient backbones used in our main experiments~\cite{xie2024sana, hacohen2024ltx}.  
These stronger backbones yield significant improvements in output quality, as reflected in the generated results.  
Notably, FLUX enables the creation of realistic yet fantastical 360$\degree$ wallpapers (\eg, nebula, asteroid, magic forest), leveraging the creative capacity of text-to-image models to depict scenes beyond the scope of existing ERP datasets.  
Video results generated with HunyuanVideo are provided in the supplementary materials.
}
\label{fig:different_backbone}
\end{figure*}

\begin{figure*}[t]
\centering
\includegraphics[width=\linewidth]{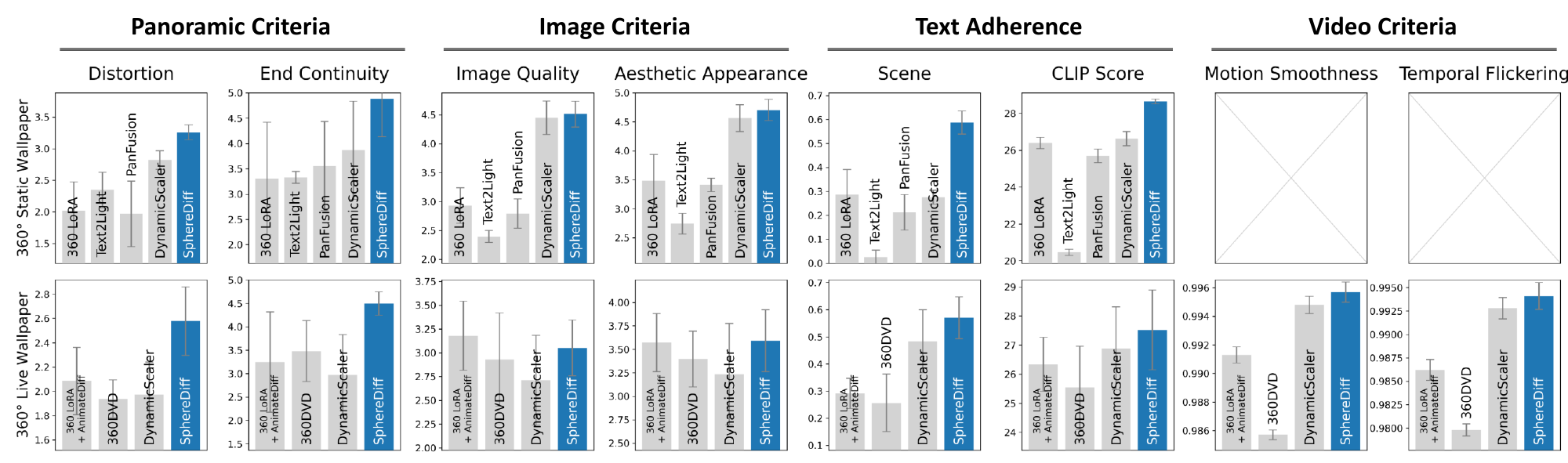}
\caption{
\textbf{Automated Quantitative Evaluation with Standard Deviations.}
We report the performance of each method across all evaluation criteria, including image, video, panoramic, and text-level aspects.
The gray line indicates the 1-sigma error bar.
The proposed method consistently surpasses the second-best method in most metrics, especially in panoramic criteria.
}
\label{fig:A_visualize_std}
\end{figure*}

\paragraph{Qualitative ERP Comparison of 360$\degree$ Live Wallpaper Generation Methods.}

In the live wallpaper comparison, a similar trend is observed.
For example, DynamicScaler~\cite{liu2024dynamicscaler} shows the same limitation, still producing stretched fireworks, as shown in the first column of \Cref{fig:A_full_erp_comparison_t2v}.
We provide video comparisons in the supplementary material.
Since 360 LoRA + AnimateDiff~\cite{guo2023animatediff} and 360DVD generate only 16 frames, we interpolate additional frames for visualization, while DynamicScaler and our method generate 121 frames.
Although 360 LoRA + AnimateDiff appears to produce reasonable 360$\degree$ live wallpapers, it struggles to animate the content meaningfully, as shown in our video.
For example, fireworks and storms lack dynamic motion; instead, they merely shift from left to right as the viewpoint changes.
360DVD also fails to generate seamless results near the poles and to faithfully reflect the input text prompts, likely due to the limited diversity in its training 360$\degree$ video dataset, as evidenced by our extensive quantitative results~(\Cref{tab:exp_user_study,tab:exp_quanti,fig:A_visualize_std}).
In contrast, our method faithfully adheres to the input text prompts and strictly respects the ERP constraints, enabled by our use of a spherical latent representation that naturally overcomes the limitations of ERP-based latent spaces.

\paragraph{More Results with Advanced Diffusion Backbones.}

Given the tuning-free nature and versatility of the proposed approach, we can apply our method to recent advanced diffusion models.
First, we apply our method to FLUX~\cite{flux2024}, a powerful text-to-image model, for 360$\degree$ static wallpaper generation.
Our approach is fully compatible with its architecture without requiring any training, enabling the creation of high-quality panoramic images.
As illustrated in \Cref{fig:different_backbone}, this combination yields outputs with superior fidelity while adhering to panoramic constraints.
To further demonstrate its versatility, we also extend our method to generate 360$\degree$ live wallpapers using HunyuanVideo~\cite{kong2024hunyuanvideo}, a state-of-the-art open-sourced text-to-video model.
This integration is also tuning-free and successfully produces visually compelling videos that preserve panoramic consistency.

\paragraph{Automated Quantitative Evaluation with Standard Deviations.}

We provide detailed automated quantitative evaluation results, including standard deviations.
As shown in \Cref{fig:A_visualize_std}, the proposed method consistently demonstrates improvements beyond the 1-sigma margin of the second-best method, statistically supporting its superiority.
Although baseline methods show comparable performance in image and video criteria, our approach significantly outperforms them in panoramic criteria and text adherence.
To ensure a fair comparison with DynamicScaler~\cite{liu2024dynamicscaler}, we do not include automated scores for 360$\degree$ wallpapers generated by advanced diffusion models~\cite{flux2024, kong2024hunyuanvideo}.
However, as qualitatively shown in \Cref{fig:different_backbone}, our method can achieve even better scores when paired with stronger base diffusion models.

\subsection{Detailed Comparison with the Other Panorama Generation Approaches}
\label{appn:comparison_baselines}

\paragraph{Data-driven 360$\degree$ Static Wallpaper Generation.}

Generating images in the Equirectangular Projection (ERP) format is challenging due to its intrinsic properties.
As previously discussed in \Cref{fig:1_motivation}, latents are unevenly distributed over the spherical surface, which leads to distortions and discontinuities when projected to the ERP domain.
For instance, latents corresponding to the topmost row in ERP must share the same value, as they represent a single spatial point (the pole).
Previous studies have attempted to address these issues through data-driven training with specialized strategies~\cite{zhang2024panfusion, feng2023diffusion360, wu2024panodiffusion, mvdiffusion, chen2022text2light} or by using alternative representations~\cite{wu2024spherediffusion, liao2023cylin, kalischek2025cubediff}.

However, high-quality 360$\degree$ panoramic images remain scarce, and most existing datasets focus on indoor scenes, such as Matterport3D~\cite{matterport3D}, Structured3D~\cite{zheng2020structured3d}, and the dataset used in MVDiffusion~\cite{mvdiffusion}, each containing approximately 10K images.
Consequently, existing methods either fail to properly adhere to the 360° constraints ~\cite{360_lora}, or struggle to adhere to the input text prompts~\cite{chen2022text2light} as shown in \Cref{fig:A_full_erp_comparison_t2i}.
This issue becomes particularly severe for applications like live wallpaper generation, which suffer from an even greater lack of data, as summarized in \Cref{tab:high_level_comparison_with_baselines}.

\paragraph{Data-driven 360$\degree$ Live Wallpaper Generation.}

The scarcity of data is even more pronounced for 360$\degree$ live wallpaper content than for static wallpapers.
For instance, 360DVD~\cite{wang2024360dvd} extends pretrained text-to-video models using an adapter, but it relies on a small dataset of merely 2,000 video clips, which is insufficient to achieve consistently high-quality results. 
To overcome the data bottleneck, GenEx~\cite{lu2024genex} captures panoramic videos from simulations to build a custom dataset. While this is an effective strategy for data augmentation, it introduces a critical constraint: the model's output is confined to the distribution of the synthetic simulation, limiting its ability to generate photorealistic content and instead producing results that appear artificial.
Our method circumvents these data dependency issues as it can be directly applied to contemporary video diffusion models.
This allows for the generation of diverse 360$\degree$ dynamic content without being constrained by the limitations of a domain-specific, often small or synthetic, training dataset.

\paragraph{Tuning-free 360$\degree$ Panorama Generation.}

To address the scarcity of 360$\degree$ video datasets, recent studies~\cite{li20244k4dgen, liu2024dynamicscaler} have attempted to generate panoramic videos using tuning-free methods. 
4K4DGen~\cite{li20244k4dgen}, for example, adopts an image-to-video pipeline by animating images generated from 360 LoRA~\cite{360_lora}.
However, this approach not only inherits the intrinsic limitations of its static image backbone but also restricts creative flexibility, as it requires a pre-existing source image for generation, as discussed in the related work section.
In contrast, direct text-to-video approaches offer greater flexibility but present different technical challenges.
DynamicScaler~\cite{liu2024dynamicscaler}, for instance, introduces an Offset Shifting Denoiser (OSD) to improve end-to-end continuity in the ERP format.
Nevertheless, because it still relies on ERP, the method produces noticeable blurry artifacts in the polar regions, as illustrated in \Cref{fig:A_full_erp_comparison_t2i,fig:A_full_erp_comparison_t2v}.

In contrast, our method addresses these issues by adopting a spherical representation.
This design not only ensures a uniform latent distribution that eliminates polar artifacts, but also facilitates the direct application of text-to-video models.
As a result, our approach enables high-quality, seamless, and tuning-free generation of both images and videos.

\begin{table}[t]
\centering
\resizebox{\columnwidth}{!}{
\begin{tabular}{l|ccccc}
\toprule
Method & Latent Space & Tuning-Free  & Wallpaper type \\
\drule
360 LoRA & ERP & \xmark & static \\
Text2Light (TOG'22) & ERP & \xmark & static \\
PanFusion (CVPR'24) & ERP & \xmark & static \\
Cubediff (ICLR'25) & Cube Map & \xmark & static \\ \midrule
360 LoRA + AnimateDiff & ERP & \xmark & live \\
360DVD (CVPR'24) & ERP & \xmark & live \\ \midrule
DynamicScaler (CVPR'25) & ERP & \cmark & static and live \\
SphereDiff (Ours) & Spherical &\cmark & static and live \\
\bottomrule
\end{tabular}
}
\caption{
\textbf{Comparison of 360$\degree$ static/live wallpaper generation approaches.}
Most existing 360$\degree$ wallpaper generation approaches perform denoising within the equirectangular latent space, whereas our method leverages spherical latents. 
Among these methods, only DynamicScaler~\cite{liu2024dynamicscaler} and ours support both static and live wallpaper generation due to their tuning-free design.
}
\label{tab:high_level_comparison_with_baselines}
\end{table}

\begin{table*}[!t]
\centering
\renewcommand{\arraystretch}{1.2}
\begin{tabular}{|p{\linewidth}|}
\hline
\textbf{Evaluation Prompt} \\  
\hline
You are an evaluator assessing an image generation model based on a single image at a time.  
Your evaluation is based on the following four criteria:  

1. \textbf{Image Quality}: Assess the overall quality of the image.  

2. \textbf{Aesthetic Appeal}: Evaluate how visually pleasing the image is.  

3. \textbf{Distortion Level}: Determine whether the image appears distorted.  
If it does not resemble a photo taken with a normal camera, it will receive a lower score.  

4. \textbf{Connectivity}: Check if the middle of the image appears disconnected.  
If there is a noticeable break, the score will be lower.  

Each criterion is rated on a five-point scale:  
Excellent (5), Good (4), Fair (3), Poor (2), and Awful (1).  

You will receive one image at a time. For each criterion, provide a concise reason for the score before listing the rating.  

Format your response as follows:  

- \textbf{Image Quality}: (Brief reason) → Score  

- \textbf{Aesthetic Appeal}: (Brief reason) → Score  

- \textbf{Distortion Level}: (Brief reason) → Score  

- \textbf{Connectivity}: (Brief reason) → Score  

\{image\}  

Please evaluate the image with the given criteria. \\  
\hline
\end{tabular}
\vspace{-0.2cm}
\caption{
\textbf{Evaluation Prompt for VLM.}
The evaluation prompt used to assess the quality of 360$\degree$ live and static wallpaper generation based on four image and panoramic criteria.
We instruct the VLM~\cite{gpt4o} to provide a score along with a brief reason to encourage chain-of-thought~\cite{wei2022chain}.
}
\label{tab:A_llm_evaluation_prompt}
\vspace{-0.1cm}
\end{table*}

\begin{figure*}[!t]
\includegraphics[width=\linewidth]{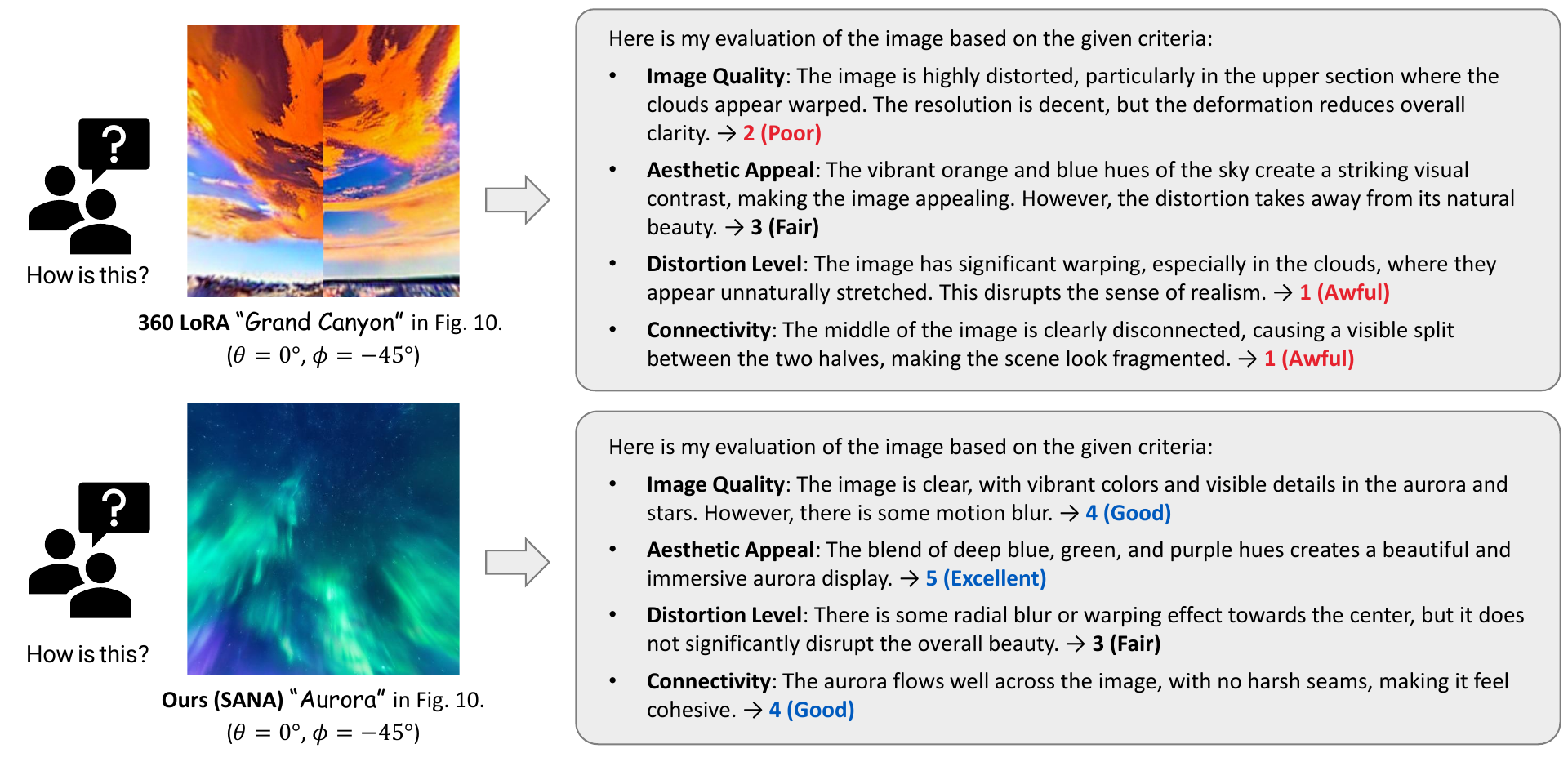}
\vspace{-0.5cm}
\caption{
\textbf{Example of VLM Responses.}  
The VLM~\cite{gpt4o} provides a concise justification along with a corresponding score.
Interestingly, we observe that the VLM demonstrates awareness of panoramic-specific artifacts such as distortion and end continuity, despite these aspects being relatively underexplored in prior VLM-based evaluation research.
}
\label{fig:A_llm_resp}
\vspace{-0.1cm}
\end{figure*}

\begin{figure*}[t]
\centering
\includegraphics[width=0.7\linewidth]{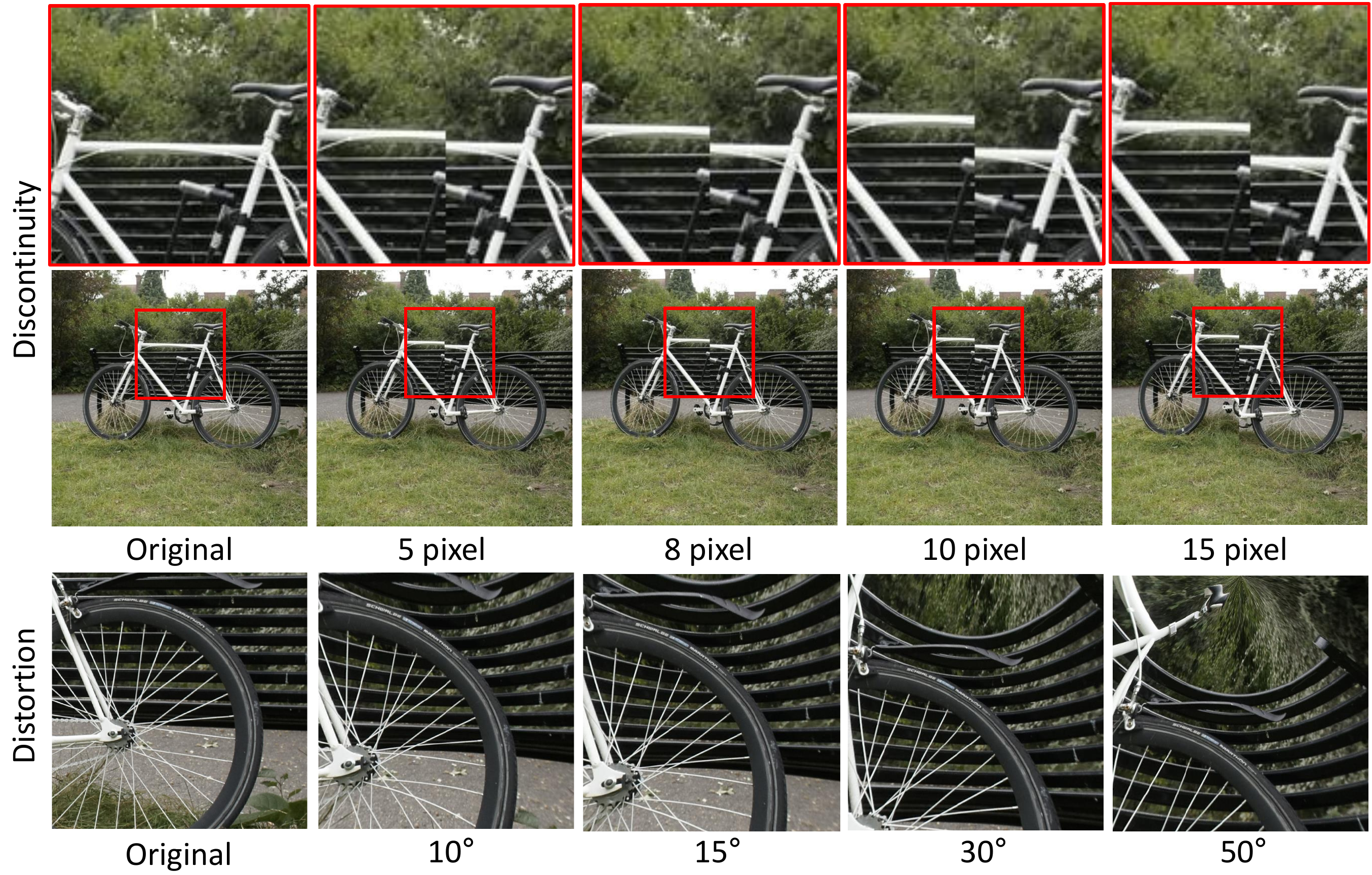}
\caption{
\textbf{Exemplars of Synthetic Distortion and Discontinuity.}
To evaluate how well the VLM~\cite{gpt4o} can recognize panoramic artifacts, we construct synthetic examples with controlled levels of distortion and discontinuity.
We vary the severity of both artifact types, as shown in the figure.
}
\label{fig:A_distorted}
\end{figure*}

\begin{figure}[t]
\centering
\includegraphics[width=\linewidth]{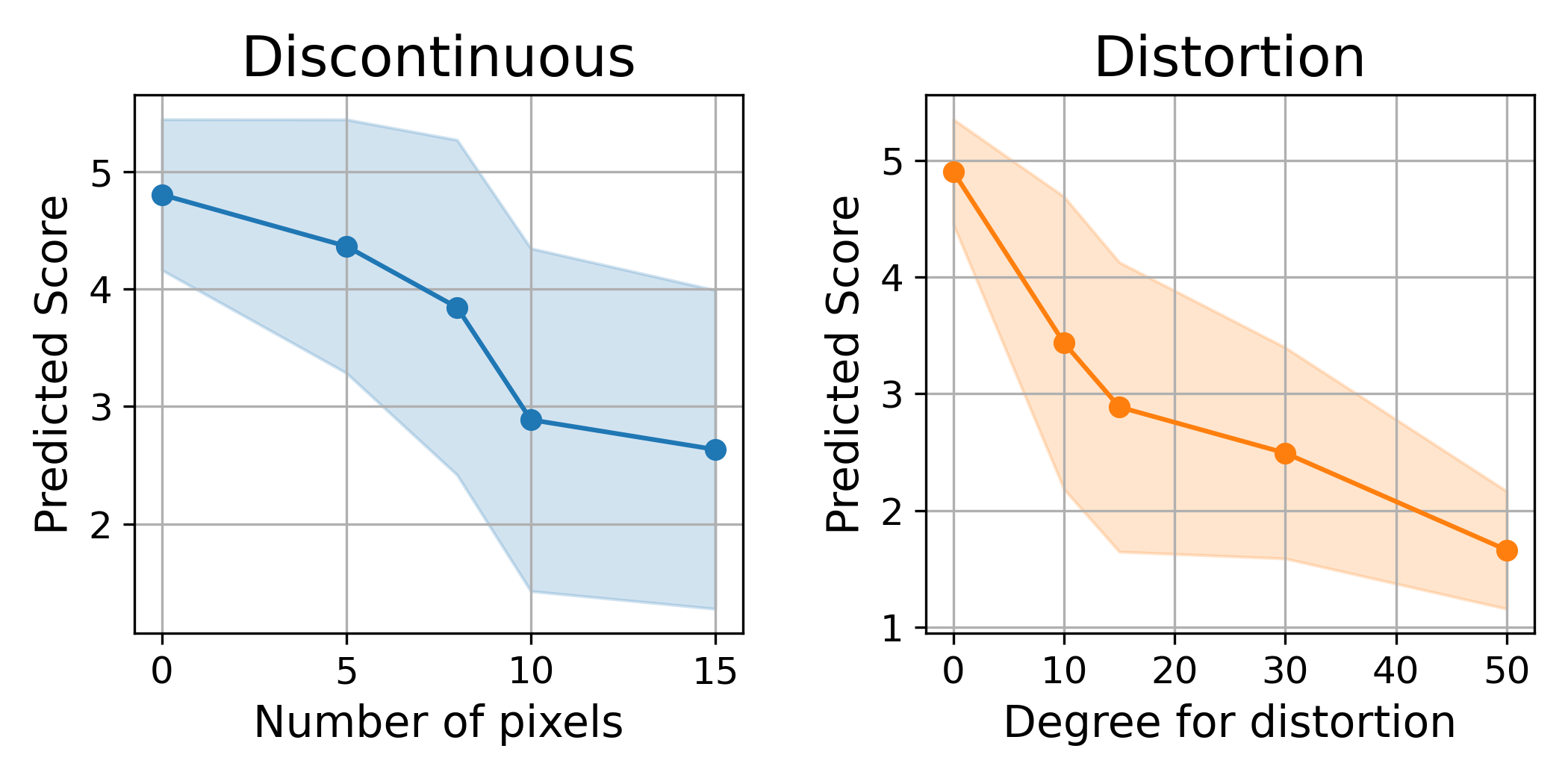}
\caption{
\textbf{Predicted Scores by GPT-4o~\cite{gpt4o} for Distortion and Continuity.}  
The scores are reported on a 1--5 scale, and the shaded area in the plot represents the 1-sigma range.  
GPT-4o demonstrates strong sensitivity to the severity of both artifact types, providing scores that closely align with the levels of distortion and discontinuity, thereby indicating the reliability of its score predictions.
}
\label{fig:llm_distort}
\end{figure}

\section{Implementation Details}
\label{appn:implementation_details}

\paragraph{Hyperparameters.}

In all MultiDiffusion setups~\cite{bar2023multidiffusion}, the base resolution and temporal length are configured to optimize the performance of each base model.
For instance, we use a resolution of $1024\times 1024$ for SANA~\cite{xie2024sana}, FLUX~\cite{flux2024}, and HunyuanVideo~\cite{kong2024hunyuanvideo}, and $512\times512$ for LTX-Video~\cite{hacohen2024ltx}.
The temporal length is fixed at 121 frames for both LTX-Video and HunyuanVideo.
For the spherical latent representation, we use 2,600 latent points on the sphere and adopt an 80$\degree$ field of view for each perspective, with 60\% overlap between neighboring views across all experiments.
The view directions are defined as follows:
\begin{itemize}
\item $\phi = \pm 90\degree$: $\theta = \frac{360\degree}{4} \cdot k,; k \in {0, 1, 2, 3}$
\item $\phi = \pm 77.5\degree$: $\theta = \frac{360\degree}{8} \cdot k,; k \in {0, \dots, 7}$
\item $\phi = \pm 45\degree$: $\theta = \frac{360\degree}{11} \cdot k,; k \in {0, \dots, 10}$
\item $\phi = \pm 22.5\degree$: $\theta = \frac{360\degree}{14} \cdot k,; k \in {0, \dots, 13}$
\item $\phi = 0\degree$: $\theta = \frac{360\degree}{15} \cdot k,; k \in {0, \dots, 14}$
\end{itemize}

This results in a total of 89 view directions.
Although we reduce the number of directions to speed up inference, this reduction can slightly compromise seamlessness due to insufficient overlap.
Nonetheless, reducing the number of views while preserving quality remains a promising direction for future research on efficient panoramic generation.

\paragraph{Tiled VAE decoding.}
To visualize the results, we decode the denoised latent representation $\mS_0$ for each view direction using a VAE decoder and stitch them together to construct an ERP image.
During this decoding process, we apply distortion-aware weighted averaging techniques to ensure seamless integration.

\paragraph{Runtime Analysis.}

Inference time per sample is approximately 3 minutes for image generation with SANA~\cite{xie2024sana} and 20 minutes for video generation with LTX-Video~\cite{hacohen2024ltx}, measured on an NVIDIA A100-40GB GPU.
Since our method does not require additional memory beyond the base requirements of the underlying models (SANA and LTX-Video), we verified that it can also run on GPUs with 24GB of VRAM.
Runtime analysis for each ablation is summarized in \Cref{tab:A_ablation}.

When using more advanced diffusion backbones, computational demands increase significantly.
Specifically, FLUX~\cite{flux2024} requires 40GB of VRAM and takes approximately 12 minutes per sample on a single H200 GPU, while HunyuanVideo~\cite{kong2024hunyuanvideo} consumes about 50GB of VRAM and requires around 3 hours for video generation.
Although our method incurs non-trivial inference time, it achieves state-of-the-art performance in both image and video quality.
Nevertheless, the relatively long runtime remains a limitation inherited from the MultiDiffusion~\cite{bar2023multidiffusion} framework.
Reducing inference time, for example by minimizing the overlapping region, presents a promising direction for future work.

\section{Evaluation Details}
\label{appn:evaluation_details}

In this section, we provide additional evaluation details that could not be included in the main paper due to space constraints.
Specifically, we present the details of the user study (\Cref{appn:evaluation_user_study}) and provide the justification for the VLM-based visual assessment (\Cref{appn:evaluation_vlm}).

\subsection{User Study Details}
\label{appn:evaluation_user_study}

We divided the 20 pairs of images and videos into five sets, each containing four pairs, allowing users to select one set for evaluation based on their convenience.  
To accurately assess distortion and end-continuity, images and videos were presented from a viewpoint with a fixed single azimuth angle ($\theta = 0\degree$) for evaluating end-continuity while varying the elevation angle ($\phi$) across five positions: $-90\degree, -45\degree, 0\degree, 45\degree, 90\degree$, as illustrated in \Cref{fig:4_quali_comparison}.
The presentation order was randomized to minimize bias.  
For image evaluation, participants selected the most suitable model from five available options, while for video evaluation, they chose from four models, selecting the one they found most appropriate for each criterion.

\balance

\subsection{VLM-based Visual Score}
\label{appn:evaluation_vlm}

\paragraph{Preprocess for Automated Evaluation.}

To automatically evaluate the 360$\degree$ static and live wallpapers, we project the ERP into perspective views to fully leverage pretrained models trained on standard perspective images.
To this end, all metrics are computed on perspective images or videos rendered with a 90$\degree$ field of view (FoV) at 14 predefined view directions.
The selected views include four azimuth angles ($0^{\circ}$, $90^{\circ}$, $180^{\circ}$, $270^{\circ}$) at each of three elevation angles ($45^{\circ}$, $0^{\circ}$, and $-45^{\circ}$), along with one view each at the top and bottom poles ($90^{\circ}$ and $-90^{\circ}$).

\paragraph{Justification of VLM-based Evaluation.}

Since tuning-free 360$\degree$ wallpaper generation methods do not rely on the source domain, and it cannot easily utilize FID-based metrics~\cite{heusel2017gans} for measuring image and panoramic criteria.
Specifically, several approaches have attempted to evaluate the quality of the generated images or videos with vision-language models~\cite{wang2025your, you2024depicting, you2024descriptive, zhou2024vision}, in terms of blur, noise, exposure, or the overall descriptive aesthetic assessment.
Thus, we utilize GPT-4o~\cite{gpt4o} for automatic evaluation for the image criteria, within the LLM-as-a-judge framework~\cite{zheng2023judging}.
\Cref{tab:A_llm_evaluation_prompt} details the prompt that we used for evaluate, and \Cref{fig:A_llm_resp} illustrates the response examples of the VLM.

\paragraph{Automated Assessment for Panoramic Criteria.}

In a manner analogous to the LLM-as-a-judge~\cite{zheng2023judging} paradigm, we also utilize GPT-4o~\cite{gpt4o} to quantify levels of distortion and discontinuity in images.
To validate the reliability of this automated evaluation, we tested GPT-4o on a curated set of 72 images exhibiting these artifacts, which were generated by randomly cropping 8 samples from 9 distinct scenes within the Mip-NeRF 360~\cite{barron2022mipnerf360} dataset.
To evaluate the reliability of panoramic quality metrics, we introduced two types of synthetic image degradation—distortion and discontinuity—at five different intensity levels, including a no-degradation case, as visually illustrated in \Cref{fig:A_distorted}.

Specifically, discontinuity is introduced by shifting the left half of the image vertically, breaking the seamless alignment.
We applied four levels of displacement: 5, 8, 10, and 15 pixels, which correspond to about 1.2\%, 1.9\%, 2.4\%, and 3.7\% of the total image height (411 pixels), respectively.
Distortion is simulated by projecting ERP to perspective views, which introduces curvature and warping artifacts.
This reflects the common distortion pattern that arises when 360$\degree$ panoramas are generated as if they were regular flat images, ignoring the spherical geometry.
The intensity levels corresponded to elevation shifts in the view direction of 10$\degree$, 15$\degree$, 30$\degree$, and 50$\degree$, respectively.

As shown in \Cref{fig:llm_distort}, our automated assessment framework demonstrates its capability to measure key panoramic criteria.
We observe a clear and systematic trend where GPT-4o~\cite{gpt4o} continuously assigns lower scores as the intensity of distortion and discontinuity increases.
This consistent, inverse relationship validates our LLM-based approach as a reliable method for quantifying such panoramic artifacts.

\section{Used Text Prompts}
\label{appn:text_prompts}

\begin{figure*}[t]
\centering
\includegraphics[width=0.98\linewidth]{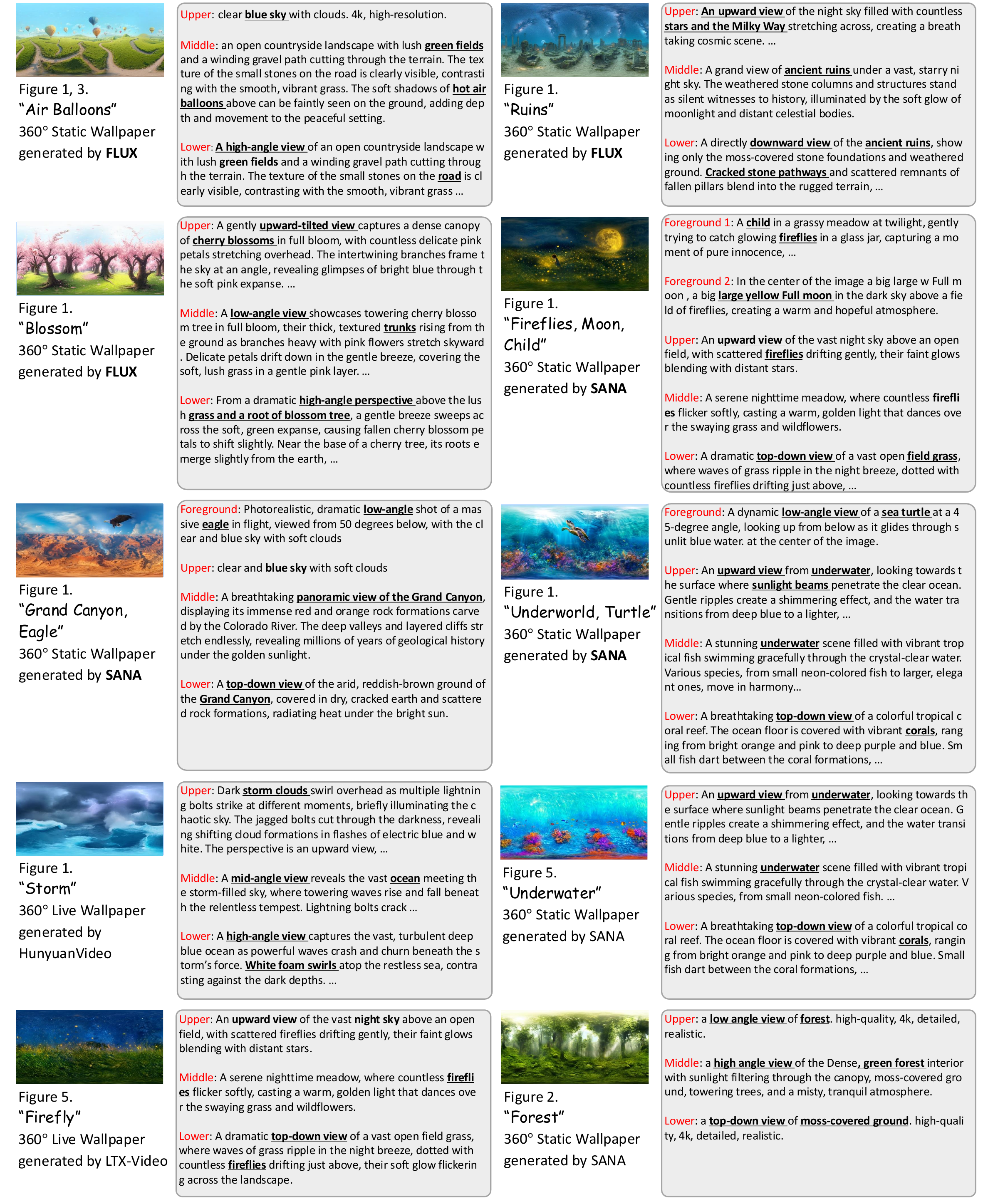}
\caption{
\textbf{Prompts Used for Generating the 360$\degree$ Static and Live Wallpapers in the Main Paper.}  
We present the text prompts corresponding to the ERP results shown in the main paper.  
Note that only \Cref{fig:1_teaser} (SANA) utilizes an additional foreground prompt; all other examples are generated without it.  
The complete list of 20 prompts will be released with the code.
}
\label{fig:A_used_prompts}
\end{figure*}

Our prompts cover 20 different scene concepts, primarily focusing on natural landscapes.
To generate 360$\degree$ live wallpapers, we use GPT-4o~\cite{gpt4o} to enrich the prompts following the prompt engineering guidelines provided on its official pages, as LTX-Video~\cite{hacohen2024ltx} struggles with very short prompts.
In contrast, we observed that SANA~\cite{xie2024sana} is relatively insensitive to prompt variations, so we use the same set of prompts for generating 360$\degree$ static wallpapers.
While the full prompt set will be publicly released, we provide several representative examples in \Cref{fig:A_used_prompts}, which were used to generate the figures in the main paper.

\clearpage

\end{document}